\documentclass[sigconf]{acmart}

\AtBeginDocument{%
  }

\setcopyright{acmlicensed}
\copyrightyear{2018}
\acmYear{2018}
\acmDOI{XXXXXXX.XXXXXXX}
\acmConference[Conference acronym 'XX]{Make sure to enter the correct
  conference title from your rights confirmation email}{June 03--05,
  2018}{Woodstock, NY}
\acmISBN{978-1-4503-XXXX-X/2018/06}

\usepackage[utf8]{inputenc} 
\usepackage[T1]{fontenc}    
\usepackage{hyperref}       
\usepackage{url}            
\usepackage{booktabs}       
\usepackage{amsfonts}       
\usepackage{nicefrac}       
\usepackage{microtype}      
\usepackage{xcolor}         
\usepackage{makecell}
\usepackage{multirow}

\usepackage{amsmath}
\usepackage{mathtools}
\usepackage{amsthm}
\usepackage{diagbox}

\usepackage{bm}
\usepackage{graphicx}
\usepackage{multirow}
\usepackage{colortbl}
\usepackage{rotating}
\usepackage{makecell}
\usepackage{wrapfig}
\usepackage{enumitem}
\usepackage{pifont}
\usepackage{makecell}
\usepackage[ruled,vlined,linesnumbered]{algorithm2e}

\usepackage[most]{tcolorbox}

\usepackage{multicol}
\usepackage{setspace}

\definecolor{best}{rgb}{1,  0.851,  0.4}
\definecolor{second}{rgb}{0.557,  0.663,  0.859}
\usepackage{hyperref}
\SetCommentSty{footnotesize}
\SetKwComment{Comment}{$\triangleright$ }{}
\SetSideCommentRight

\begin{document}

\title{$\mathtt{GDformer}$: Going Beyond Subsequence Isolation for Multivariate Time Series Anomaly Detection}

\author{Qingxiang Liu}
\authornote{Email: qingxiangliu737@gmail.com}
\affiliation{%
  \institution{The Hong Kong University of Science and Technology (Guangzhou)}
  \country{China}
}

\author{Xiaoliang Luo}
\affiliation{%
  \institution{China Mobile (Jiangxi) Virtual Reality Technology Co., Ltd.}
  \country{China}
}

\author{Chenghao Liu}
\affiliation{%
  \institution{Salesforce AI Research}
  \country{USA}
}

\author{Sheng Sun}
\affiliation{%
  \institution{Institute of Computing Technology, Chinese Academy of Sciences}
  \city{Beijing}
  \country{China}
}

\author{Di Yao}
\affiliation{%
  \institution{Institute of Computing Technology, Chinese Academy of Sciences}
  \city{Beijing}
  \country{China}
}

\author{Lvchun Wang}
\affiliation{%
  \institution{China Mobile (Jiangxi) Virtual Reality Technology Co., Ltd.}
  \country{China}
}

\author{Wei Yu}
\affiliation{%
  \institution{China Mobile (Jiangxi) Virtual Reality Technology Co., Ltd.}
  \country{China}
}

\author{Yuxuan Liang}
\authornote{Corresponding Author. Email: yuxliang@outlook.com}
\affiliation{%
  \institution{The Hong Kong University of Science and Technology (Guangzhou)}
  \country{China}
}


\begin{abstract}
Multivariate Time Series (MTS) anomaly detection is crucial for maintaining the reliability of real-world systems, which can identify malfunctions from operational data and mitigate the potential financial losses. 
Given the scarcity of anomalies and difficulty in labeling MTS data, the MTS anomaly detection task is usually framed as an unsupervised learning problem to derive a compact and distinguished detection criterion without accessing the anomaly points.
The existing deep learning methods are all confined to \textit{isolated subsequences} with limited and subsequence-contained horizons, hardly deriving anomaly scores with series-level knowledge and promising the unified detection criterion.
In this paper, we propose the \textbf{G}lobal \textbf{D}ictionary-enhanced Trans\textbf{former} ($\mathtt{GDformer}$) with a renovated dictionary-based cross attention mechanism to cultivate the global representations shared by all normal points in the entire series. 
Accordingly, the cross-attention maps reflect the correlation weights between the point and global representations, which naturally leads to the representation-wise \textit{similarity}-based detection criterion.
To foster more compact detection boundary, prototypes are introduced to capture the distribution of normal point-global correlation weights.
Extensive experimental results validate the effectiveness and efficiency of $\mathtt{GDformer}$, compared with state-of-the-art unsupervised anomaly detection methods on six popular benchmark datasets.
\end{abstract}



\received{20 February 2007}
\received[revised]{12 March 2009}
\received[accepted]{5 June 2009}

\maketitle

\begin{figure*}[!t]
	\centering
	\includegraphics[width=\textwidth]{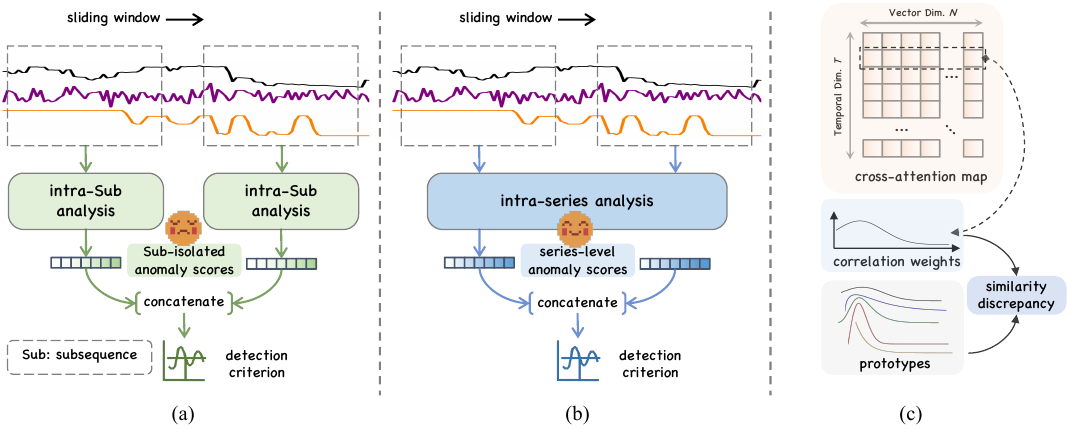}
	\caption{Difference in deriving the detection criterion.
    (a) Existing methods learn intra-subsequence temporal knowledge and derive the detection criterion by combining subsequence-level anomaly scores.
    (b) Our proposal cultivates global normal knowledge to provide series-level criterion.
    (c) The point-global correlation manifests as each row of the cross-attention map, based on which prototypes are cultivated to evaluate similarity discrepancy.
    }
    \label{motivation}
\end{figure*}

\begin{figure}[!t]
	\centering
	\includegraphics[width=\columnwidth]{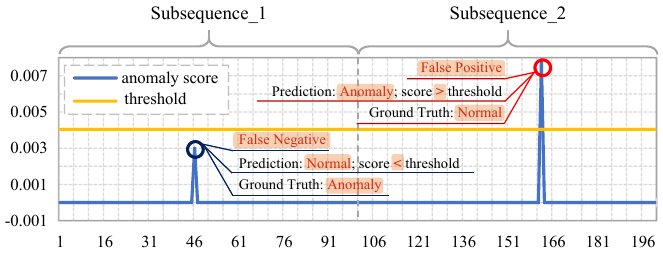}
	\caption{Anomaly scores v.s. detection threshold for different subsequences in AnomalyTrans \cite{xu2022anomaly}.}
	\label{motivation_example}
\end{figure}

\section{Introduction}

Many real-world systems usually encompass multiple interrelated sensors for different measurements. For example, in a greenhouse control system, multi-sensors monitor the temperature, humidity, light intensity, etc., for further intelligent maintenance. 
With these systems running consecutively, large-scale time series of multi-dimensional observations can be generated and then extensively analyzed for identifying the normal work mode to further detect malfunctions which manifest as anomalous observations~\cite{li2021block,wen2022kddtimeseries,yang2023sgdpstreamgraphneuralnetwork}.
This is of great value to ensuring system security and reducing financial losses.
Given its importance, many methods for multivariate time series (MTS) anomaly detection have been proposed, among which the unsupervised ones are paid more attention to, due to the rarity of anomalous time points and the difficulty of labeling multi-dimensional time series data~\cite{9373923,zhang2018deepneuralnetworkunsupervised,zhao2020multivariatetimeseriesanomalydetection,Zhang_2022}. Therefore, we also delve into unsupervised time series anomaly detection.

In unsupervised setting, different pretext tasks are devised to learn the shared representations among normal time points, which are deemed to distinguish from abnormal representations.
Generally speaking, the existing deep learning methods can be categorized into two groups~\cite{sánchezferrera2025reviewselfsupervisedlearningtime,qiu2025tab}, i.e., \textit{forecasting}- and \textit{reconstruction}-based ones.
In forecasting-based methods, the future values are predicted based on lookback windows and the prediction models are trained by minimizing prediction errors between ground truth and predicted values~\cite{yu2025merlin,wu2024multi,liu2024wftnet}. The potential anomalous points will be detected if the prediction errors exceed the defined criterion. 
However, these methods are susceptible to segment anomalies, where the prediction errors may be lowered if anomaly points concurrently exist in inputs.
In reconstruction-based methods, the reconstruction errors of anomalies are higher than those of normal time points, due to the well-cultivated temporal representations in the training process~\cite{10.5555/3618408.3619209,yang2023dcdetector,xu2022anomaly,yang2023dcdetector,xie2025multivariate}.
However, given the rarity of anomalies and complex temporal patterns, the decision criterion may be dominated by normal points, thus leading to poor distinguishability.


As shown in Fig. \ref{motivation} (a), these methods follow such pipeline to obtain the detection criterion: 
(i) dividing the entire \textit{series} into non-overlapped \textit{subsequences} (which can be seen as \textit{samples} in deep learning); (ii) learning intra-subsequence temporal knowledge and isolated anomaly scores; (iii) determining the unified detection criterion for all points in whatever subsequences.
Therefore, such \textit{subsequence isolation} approach focuses on limited horizon which is much less context-informative compared with the entire series.
Moreover, given the \textit{heterogeneity} across subsequences in terms of temporal fluctuation and the number of anomalies, the derived point-wise anomaly scores are highly subsequence-contained.
As shown in Fig. \ref{motivation_example}, directly concatenating such anomaly scores to derive the global detection criterion for the entire series results in false negative and false positive cases.

A prospective approach is to enlarge the horizon to the entire series so as to cultivate global representations shared by all normal points, which further ensures the series-level anomaly scores and detection criterion for any points (Fig. \ref{motivation} (b)). 
However, it is nontrivial to learn such global representations and then derive unified anomaly scores, given the following challenges.
(i) The self-attention mechanism in Transformers has the poor $\mathcal{O}(n^2)$ time and space complexity, with $n$ denoting the number of tokens. 
Therefore, directly inputting the entire series, with each time point corresponding to a token, will lead to the enormous scale of attention maps, which lags the training process and challenges the memory size.
(ii) Supposing we obtain the well-cultivated global representations, a natural detection criterion is that the \textit{similarity discrepancy} of global-abnormal representations is higher than that of the global-normal ones.
Given the numerous and complex temporal representations in the entire series, the simple statistical approaches, i.e., Kullback–Leibler (KL) divergence \cite{6832827} and Jensen-Shannon (JS) divergence \cite{1365067} are incompetent in evaluating the similarity between the inherent temporal patterns (see \underline{Sec.~\ref{similarity_ablation}}).

To address these challenges, we propose a \textit{similarity-based} anomaly detection method, which augments the Transformer with the \textit{global dictionary} of Key and Value vectors to provide global discrete latent representations shared by all normal points. 
Therefore, after the cross attention operation between Query and Key vectors, each row in the cross-attention maps can represent the correlation weights between a given point and global representations (Fig. \ref{motivation} (c)).
Moreover, due to the much smaller size of the global dictionary, the computational and memory efficiency will be improved in attention process (see \underline{ Sec.~\ref{model_efficiency}}).
For the second challenge, we introduce \textit{prototypes} to capture the normal distribution of cross-attention weights.
Therefore, well-cultivated prototypes have higher discrepancy with the abnormal point-global correlation weights, promising an effective anomaly detection criterion.
We term our model the \textbf{G}lobal \textbf{D}ictionary-enhanced Trans\textbf{former} ($\mathtt{GDformer}$).
The contributions of the paper can be summarized as follows.
\begin{itemize}[leftmargin=*]
\item We propose $\mathtt{GDformer}$ with a global dictionary of Key and Value vectors to learn global representations shared by all normal points in the entire series, which alleviates the effects of subsequence isolation and ensures the true unified detection criterion.
\item We introduce prototypes to capture normal point-global correlation patterns, which enables distinguishable normal-abnormal similarity discrepancy and provides compact decision boundary.
\item Extensive experiments on six popular benchmark datasets validate the effectiveness and efficiency of $\mathtt{GDformer}$. 
\end{itemize}


\begin{figure*}[!t]
	\centering
	\includegraphics[width=\textwidth]{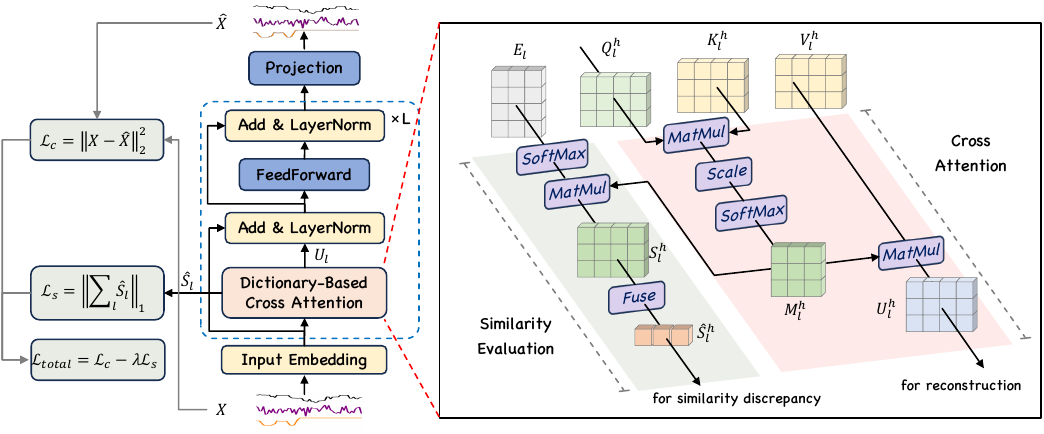}
	\caption{The overall architecture of $\mathtt{GDformer}$. 
    The \textbf{right} shows the dictionary-based cross attention mechanism in each head $h$, including the modules of cross attention (see \underline{Sec. \ref{cross_attention}}) and similarity evaluation (see \underline{Sec. \ref{similarity_evaluation}}).
    }
	\label{method}
\end{figure*}

\section{Related Work}
\textbf{Statistical methods} learn the statistical characteristics of time series data~\cite{lu2008network,mahimkar2011rapid,rasheed2009fourier,siffer2017anomaly}. For example, ~\cite{rasheed2009fourier} identifies high frequency changes and outlier regions by adopting fast fourier transform. In~\cite{siffer2017anomaly}, outliers in stream are detected based on extreme value theory. These methods are computation-lightweight but non-effective for complex MTS anomaly detection.

\textbf{Machine learning methods} include clustering-based, density-based, and classification-based ones. 
In clustering-based methods, the distance to clustering centers is termed as the anomaly score. 
SVDD~\cite{tax2004support} and Deep-SVDD~\cite{ruff2018deep} can enable the compact cluster for normal representations. THOC~\cite{shen2020timeseries} fuses multi-resolution features by a hierarchical clustering mechanism and identifies abnormal features by multiple distances. ITAD~\cite{shin2020itad} conducts cluster analysis on decomposed tensor and calculates distances as anomaly scores.
In density-based methods, the density of temporal representations is calculated for outlier determination, including LOF~\cite{breunig2000lof}, COF~\cite{tang2002enhancing}, DAGMM \cite{zong2018deep}, and MPPCACD \cite{yairi2017data}.
The classification-based methods treat time series anomaly detection as a classification task and accordingly employ the classification methods, such as decision trees \cite{khraisat2018anomaly}, SVM \cite{hu2003robust}, and OCSVM~\cite{amer2013enhancing} for anomaly detection.

\textbf{Deep learning methods} are roughly divided into forecasting-based and reconstruction-based ones. 
In the former, future values are predicted and forecasting errors are formalized as the anomaly scores~\cite{sánchezferrera2025reviewselfsupervisedlearningtime,qiu2025tab}. The representative methods include LSTM~\cite{hundman2018detecting} and GANs~\cite{yao2022kfreqgan}.
Reconstruction-based methods involve reconstructing the input time series and reconstruction errors are termed as anomaly scores~\cite{park2018multimodal,su2019robust,zhou2019beatgan}.
Another line of reconstruction-based methods does not directly employ reconstruction errors as anomaly scores~\cite{song2023memto,xiao2023imputation,he2025graph,xu2022anomaly,yang2023dcdetector,luo2024moderntcn,feng2024sensitivehue,xie2025multivariate}. For example, Anomaly Transformer (AnomalyTrans) embodies a novel anomaly-attention mechanism to learn point-wise series- and prior-association and then derives the association discrepancy-based criterion \cite{xu2022anomaly}.

In these methods, Transformer \cite{NIPS2017_3f5ee243} is widely used to learn temporal representations, due to its effectiveness in modeling sequential data. 
However, the receptive fields of Transformer largely depend on the horizons of input subsequences, resulting in less context-informative representations and inconsistent detection criterion for subsequence-specific points. 
By contrast, we propose the $\mathtt{GDformer}$, which goes beyond subsequence isolation strategy via the introduction of a dictionary-based cross-attention mechanism to cultivate global normal representations with series-level context information and derive the unified similarity-based criterion.

\section{Methodology}
\label{Sec3}
Supposing there are $d$ devices in a real-world system, the observations in the duration of $\mathcal{T}$ can be denoted as a time series $\boldsymbol{\mathcal{X}} = (\boldsymbol{x}_1, \boldsymbol{x}_2, \dots, \boldsymbol{x}_\mathcal{T})$, where $\boldsymbol{x}_t \in \mathbb{R}^d$ represents these $d$ measurements at time $t$.
The aim of MTS anomaly detection is to determine whether the observation at time $t$ is anomalous or not, i.e., yielding $\boldsymbol{\mathcal{Y}} = (y_1, y_2, \dots, y_{\mathcal{T}})$, where $y_t=1$ if $\boldsymbol{x}_t$ is anomalous and $y_t=0$ otherwise.
In the training process, the whole time series are usually divided into non-overlapped subsequences with $T$ time steps and then input into the designed models for representation learning~\cite{xu2022anomaly,yang2023dcdetector}. Without loss of generality, we denote $\boldsymbol{X} \in \mathbb{R}^{T\times d}$ as a subsequence.

\subsection{Model Architecture}
The overall structure of our proposed $\mathtt{GDformer}$ is shown in Fig. \ref{method}. Overall, $\mathtt{GDformer}$ stacks the dictionary-based cross-attention module and feed-forward layers alternatively for representation learning, with a projection block for reconstruction. 
For the input subsequence $\boldsymbol{X}\in \mathbb{R}^{T\times d}$, we randomly mask the $T\times d$ observation values with the probability of $\alpha$.
Then, the masked subsequence is normalized via instance normalization \cite{kim2022reversible,Ulyanov_2017_CVPR}, denoted as $\tilde{\boldsymbol{X}}\in \mathbb{R}^{T\times d}$, to mitigate the effects of observation noise. 
We adopt a linear layer to obtain $\tilde{\boldsymbol{X}} \in \mathbb{R}^{T\times D}$, where $D$ is the target dimension.
Each point $\tilde{\boldsymbol{x}}_t \in \mathbb{R}^D$ in $\tilde{\boldsymbol{X}}$ is termed as a temporal token. The input embedding of the first layer $\boldsymbol{X}_0 =  \tilde{\boldsymbol{X}}$.
The overall operations of the $l$-th layer $(l\in[1,L])$ can be formulated as:
\begin{equation}
\begin{array}{l}
\boldsymbol{U}_l = \text{LN}(\boldsymbol{X}_{l-1} + \text{CA}(\boldsymbol{X}_{l-1}, \boldsymbol{K}_l, \boldsymbol{V}_l)), \\
\boldsymbol{X}_l = \text{LN}(\boldsymbol{U}_l + \text{FeedForward}(\boldsymbol{U}_l)),
\end{array}
\end{equation}
where LN represents the layer normalization and CA represents our proposed dictionary-based cross attention mechanism. $\boldsymbol{K}_l$ and $\boldsymbol{V}_l$ denote the learnable Key and Value vectors in the global dictionary. $\boldsymbol{U}_l$ denotes the hidden temporal representation. The representation $\boldsymbol{X}_L$ from the $L$-th layer is input into a linear projection layer for reconstruction.
We elaborate the details of dictionary-based cross attention in the following part.


\subsection{Dictionary-Based Cross Attention}
\label{DBCA}
The canonical Transformers learn the correlation of different temporal tokens via self-attention mechanism, where the triple inputs, i.e., Query, Key, and Value are all derived by the linear projection of $\boldsymbol{X}_l$ \cite{NIPS2017_3f5ee243}.
Compared with the entire series, the subsequence is constrained to limited horizons and is less context-informative.
Therefore, the cultivated temporal representations from the self-attention mechanism can only learn intra-subsequence knowledge.
On the other hand, given the heterogeneity of abnormal points and temporal distribution in different subsequences, the subsequence-isolated analysis can hardly ensure the global detection criterion for all points.
Hence, in this section, we devise a novel dictionary-based cross attention mechanism to foster series-level global representations shared by normal points, which naturally guarantees the unified anomaly evaluation and detection criterion.

\subsubsection{Cross Attention.} 
\label{cross_attention}
We denote $\boldsymbol{K}_l \in \mathbb{R}^{N\times D}$ and $\boldsymbol{V}_l \in \mathbb{R}^{N\times D}$ as the $N$ Key and Value vectors of the global dictionary in the $l$-th Transformer layer.
In dictionary-based cross attention block, for each head $h\in {1, 2, \cdots, H}$, we define the Query matrix $\boldsymbol{Q}_l^h = \boldsymbol{X}_{l-1} W_l^h$, where $W_l^h \in \mathbb{R}^{D\times D_h}$ and $D_h = \lfloor \frac{D}{H} \rfloor$.
Note that we directly split $\boldsymbol{K}_l$ and $\boldsymbol{V}_l$ into $[\boldsymbol{K}_l^h \in \mathbb{R}^{N\times D_h} ]_1^H$ and $[\boldsymbol{V}_l^h \in \mathbb{R}^{N\times D_h}]_1^H$ for each head $h$, instead of using linear projection layers.
Then, the operation of cross-attention in head $h$ can be defined as:
\begin{equation}
    \boldsymbol{U}_l^h = {\rm Softmax}(\frac{\boldsymbol{Q}_l^h\boldsymbol{K}_l^{h\top}}{\sqrt{D_h}})\boldsymbol{V}_l^h.
\end{equation}
We can fuse $\boldsymbol{U}_l^h \in \mathbb{R}^{T\times D_h}$ in each head to obtain $\boldsymbol{U}_l \in \mathbb{R}^{T\times D}$ for reconstruction.
In the unsupervised training process, $\boldsymbol{K}_l$ and $\boldsymbol{V}_l$ are updated iteratively with all temporal points, which can learn the shared representations of normal points in the entire series.

\subsubsection{Similarity Evaluation.} 
\label{similarity_evaluation}
Let $\boldsymbol{M}_l^h = {\rm Softmax}(\frac{\boldsymbol{Q}_l^h\boldsymbol{K}_l^{h\top}}{\sqrt{D_h}})$ denote the cross-attention map.
Each row in $\boldsymbol{M}_l^h \in \mathbb{R}^{T\times N}$ reflects the distribution of correlation weights between each temporal representation and the global representation $\boldsymbol{K}_l^h$.
We can directly compare the discrepancy of such distribution and then determine a detection criterion.
However, the conventional methods mainly adopt statistics methods such as KL divergence and JS divergence to evaluate distribution difference, instead of the similarity discrepancy between representations from the perspective of the inherent temporal patterns.
Therefore, the strategy fails to guarantee the homogeneity between numerous temporal representations and the global representations, hardly leading to compact decision boundary (comparison results in \underline{Sec.~\ref{similarity_ablation}}).
Therefore, in our devised dictionary-based cross attention mechanism, we introduce an extra branch for similarity evaluation.

Let $\boldsymbol{E}_l \in \mathbb{R}^{P\times N}$ denote $P$ prototypes in the $l$-th layer to capture the distribution patterns of normal-global correlation weights.
Then, we calculate the similarity between $\boldsymbol{E}_l$ and $\boldsymbol{M}_l^h$ as:
\begin{equation}
    \boldsymbol{S}_l^h = \boldsymbol{M}_l^h {\rm Softmax}(\boldsymbol{E}_l)^\top,
    \label{sim_eval}
\end{equation}
where we first conduct row-wise normalization of the prototypes $\boldsymbol{E}_l$ via Softmax.
Each row in $\boldsymbol{S}_l^h \in \mathbb{R}^{T\times P}$ represents the similarity between the point-global correlation distribution (in $\boldsymbol{M}_l^h$) and the prototypical distribution patterns ${\rm Softmax}(\boldsymbol{E}_l)$.
We then fuse $\boldsymbol{S}_l^h$ by row to obtain $\hat{\boldsymbol{S}}_l^h \in \mathbb{R}^T$ and then calculate the summation over heads to obtain $\hat{S}_l \in \mathbb{R}^T$, where each scalar represents the similarity strength of the corresponding point.
Higher values reflect stronger similarity between the correlation weights and prototypes, naturally promising a similarity-based criterion. 
Detailed process of dictionary-based cross-attention mechanism, including cross attention and similarity evaluation, is presented in Algorithm \ref{ag2}.

\begin{algorithm}[!ht]
    \caption{Dictionary-based Cross Attention}
    \label{ag2}
    \KwIn{$\boldsymbol{X}_{l-1} \in \mathbb{R}^{T\times D}$;  
    $\boldsymbol{K}_l^h \in \mathbb{R}^{N\times D_h}$, $\boldsymbol{V}_l^h \in \mathbb{R}^{N\times D_h}$ ($h \in [1,H]$), and $\boldsymbol{E}_l \in \mathbb{R}^{P\times N}$}
    \For{$h \in [1, H]$}{
        $\boldsymbol{Q}_l^h=\boldsymbol{X}_{l-1}\boldsymbol{W}_l^h $ \Comment*[r]{$\boldsymbol{W}_l^h \in \mathbb{R}^{D\times D_h}, \boldsymbol{Q}_l^h \in \mathbb{R}^{T\times D_h} $}
        
        $\boldsymbol{M}_l^h = {\rm Softmax}(\frac{\boldsymbol{Q}_l^h\boldsymbol{K}_l^{h\top}}{\sqrt{D_h}})$
        \Comment*[r]{$\boldsymbol{M}_l^h \in \mathbb{R}^{T\times N}$}
        
        $\boldsymbol{U}_l^h = \boldsymbol{M}_l^h \boldsymbol{V}_l^h$ 
        \Comment*[r]{$\boldsymbol{U}_l^h \in \mathbb{R}^{T\times D_h}$}
        
        $\boldsymbol{S}_l^h = \boldsymbol{M}_l^h {\rm Softmax}(\boldsymbol{E}_l)^\top$ \Comment*[r]{$\boldsymbol{S}_l^h \in \mathbb{R}^{T\times P}$s}
        
        $\hat{\boldsymbol{S}}_l^h = {\rm Sum}(\boldsymbol{S}_l^h, {\rm dim}=1)$ \Comment*[r]{$\hat{\boldsymbol{S}}_l^h \in \mathbb{R}^T$}
        
    }
    $\boldsymbol{U}_l={\rm Concat}([\boldsymbol{U}_l^1, \cdots, \boldsymbol{U}_l^H], {\rm dim}=1)$ \Comment*[r]{$\boldsymbol{U}_l \in \mathbb{R}^{T \times D}$}
    
    $\hat{\boldsymbol{S}}_l = {\rm Sum}([\hat{\boldsymbol{S}}_l^1, \cdots, \hat{\boldsymbol{S}}_l^H], {\rm dim}=1)$ \Comment*[r]{$\hat{\boldsymbol{S}}_l \in \mathbb{R}^{T}$}
    
    \Return $\boldsymbol{U}_l$ and $\hat{\boldsymbol{S}}_l$ \Comment*[r]{$\boldsymbol{U}_l$ for reconstruction; $\hat{\boldsymbol{S}}_l$ for similarity discrepancy}
\end{algorithm}

\subsection{Training and Inference}

We adopt the reconstruction loss ($\mathcal{L}_c$) to guide the global dictionary to learn the shared representations of the series-level normal points.
To further guarantee prototypes learn the normal distribution patterns, the similarity discrepancy ($\mathcal{L}_s$) between the cross-attention weights and prototypes is introduced, which can be formulated as:
\begin{align}
{\mathcal{L}_{total}} &= {\mathcal{L}_c} - \lambda {\mathcal{L}_s} = \left\| {\boldsymbol{X} - \hat{\boldsymbol{X}}} \right\|_2^2 - \lambda {\left\| {\sum\nolimits_l { {\hat{\boldsymbol{S}}_l} } } \right\|_1},
 \label{loss_total}
\end{align}
where $\hat{\boldsymbol{X}} \in \mathbb{R}^{T\times d}$ denotes the reconstruction results of $\boldsymbol{X}$. 
$\| \cdot \|_*$ denotes the $*$-norm. $\lambda$ is adopted to balance the two loss items. We set $\lambda > 0$ to enlarge the similarity degree between the prototypes and the cross-attention weights in unsupervised learning.
We can observe that $\mathcal{L}_s$ sums the similarity values in all $L$ layers and can aggregate multi-scale distribution knowledge, thereby leading to an informative measure.

\textbf{Anomaly Detection Threshold.}
After optimization, the prototypes can learn the distribution of the correlation weights between normal temporal representations and the global representations. 
Therefore, the similarity discrepancy of abnormal attention weights and prototypes is higher than that of normal ones. Naturally, in inference process, we can obtain the anomaly score of $\boldsymbol{X} \in \mathbb{R}^{T\times d}$ as:
\begin{equation}
    {\rm AnomalyScore}(\boldsymbol{X}) = {\rm Softmax} \left( -{\sum\nolimits_l { {\hat{\boldsymbol{S}}_l} } } \right),
\end{equation}
where ${\rm AnomalyScore}(\boldsymbol{X}) \in \mathbb{R}^T$ indicates the point-wise anomaly scores for $T$ points and has higher values for abnormal points.
Let $\delta$ denote the series-level anomaly threshold. We can obtain the detection output $\boldsymbol{\mathcal{Y}}$ as:
\begin{equation}
 \boldsymbol{\mathcal{Y}} = \{y_i\}_1^{\mathcal{T}}, {y_i}= \left\{ {\begin{array}{*{20}{c}}
{1,}&{ {\rm AnomalyScore}{{(\boldsymbol{x}_i)}} \ge \delta ,}\\
{0,}&{ {\rm AnomalyScore}{{(\boldsymbol{x}_i)}} < \delta .}
\end{array}} \right.
\end{equation}

\subsection{Analysis and Discussion}
\textbf{Transferability.}
As for the research of foundation models in time series analysis, one can train a unified model for cross-domain time series datasets~\cite{liang2024foundation}. 
Furthermore, we have a key observation that the global dictionary have great transferability (details in Sec.~\ref{Transferability}), which validates that cross-domain datasets may have the shared normal temporal patterns, thus laying the foundation for the construction of time series anomaly detection foundation model. 

\textbf{Time and Space Complexity.}
The time complexity of cross attention is $\mathcal{O}(TDN)$, where $T$, $D$, and $N$ represent the number of temporal tokens, the Transformer dimension, and dictionary size respectively. The time complexity of similarity evaluation is $\mathcal{O}(TNP)$, where $P$ represents the number of prototypes. Therefore, the time complexity of the novely-proposed dictionary-based cross attention is $\mathcal{O}(TDN + TNP)$.

The space complexity of the cross-attention map $M$, the cross-attention results $U$, and the similarity $S$ is $\mathcal{O}(TN), \mathcal{O}(TD),$ and $\mathcal{O}(TP)$ respectively. Hence, the space complexity of the dictionary-based cross attention is $\mathcal{O}(TN+TD+TP)$.

\textbf{Complexity Comparison.}
Given the same model settings with the Transformer dimension denoted as $D$ and $H$ heads, the number of parameters in one Anomaly-Attention layer \cite{xu2022anomaly} is formulated as :
\begin{equation}
    3D \times ({D_h} \times H) + D \times H,
    \label{num_an}
\end{equation}
where the first addend corresponds to multi-head self-attention and the second to the prior-association.
The number of parameters in the one attention layer in DCdetector \cite{yang2023dcdetector} is formuated as:
\begin{equation}
    3D \times ({D_h} \times H).
    \label{num_dc}
\end{equation}

We can obtain the parameter amount of the dictionary-based cross-attention as:
\begin{equation}
    D \times ({D_h} \times H) + 2N \times ({D_h} \times H) + P \times N,
    \label{num_mem}
\end{equation}
where the first addend correspond to the input projection for Query and the second to the learnable Key-Value matrices and the last to the prototypes.

In our implementation, we have the number of prototypes $P$ and the dictionary size $N$ much less than the Transformer dimension $D$, i.e., $P\ll D, N \ll D$. Therefore, we can obtain the following derivations:
\begin{align}
    & P \times N \ll D \times N < D \times (D - N) \notag \\
    & < 2D \times (D - N) = 2(D - N) \times {D_h} \times H.
\end{align}
Therefore, 
\begin{equation}
    P \times N + 2N \times {D_h} \times H \ll 2D \times {D_h} \times H.
\end{equation}
Finally, we can obtain
\begin{equation}
    D \times {D_h} \times H + 2N \times ({D_h} \times H) + P \times N \ll 3D \times {D_h} \times H.
\end{equation}
That is, given the same parameter settings of the attention layer, the parameter amount of $\mathtt{GDformer}$ is much less than those of AnomalyTrans and DCdetector. Detailed comparison results in model efficiency can be found in \underline{Sec. \ref{model_efficiency}}.



\begin{table}[!tbp]
  \small
  \centering
  \tabcolsep=0.12cm
  \renewcommand\arraystretch{1.2}
  \caption{The details and hyperparameter settings of the adopted datasets. AR: abnormal proportion.}
    \begin{tabular}{c|ccccc|ccccc}
    \toprule
    Dataset  & \#Entities & $d$ & \#Training & \#Test & AR (\%) & $\lambda$     & $P$     & $N$     & $\delta$  \\
    \midrule
    MSL   &27 & 55      & 58,317 & 73,729 & 10.5  & 3     & 12    & 16    & 0.8  \\
    SMAP  &55 & 25      & 135,183 & 427,617 & 12.8  & 2     & 12    & 6     & 0.7\\
    SWaT  & 1 & 51      & 496,800 & 449,919 & 12.1 & 2     & 8     & 8     & 0.5 \\
    PSM  & 1 & 25      & 55,541  & 87,841 & 27.8  & 1     & 10    & 10    & 0.6\\
    GECCO & 1 & 9       & 69,260  & 69,260 & 1.05  & 4 & 10 & 10 & 0.5\\
    ASD    & 12 & 19      & 102,331   & 51,840 & 4.61 & 4 & 6 & 12 & 0.5\\
    \bottomrule
    \end{tabular}%
  \label{details}%
\end{table}%

\begin{table*}[!tbp]
  \centering
  \caption{Anomaly detection performance comparisons in point-adjustment metrics. \textcolor{red}{\textbf{Bold}}: the best. \textcolor{blue}{\underline{Underline}}: the second best. }
    \begin{tabular}{c|ccc|ccc|ccc|ccc}
    \toprule
    Dataset  & \multicolumn{3}{c|}{MSL} & \multicolumn{3}{c|}{SMAP} & \multicolumn{3}{c|}{SWaT} & \multicolumn{3}{c}{PSM} \\
    \midrule
    Metric    & P     & R     & F1    & P     & R     & F1    & P     & R     & F1    & P     & R     & F1 \\
    \midrule
    OCSVM   & 59.78  & 86.87  & 70.82  & 53.85  & 59.07  & 56.34  & 45.39  & 49.22  & 47.23  & 62.75  & 80.89  & 70.67  \\
    IForest   & 53.94  & 86.54  & 66.45  & 52.39  & 59.07  & 55.53  & 49.29  & 44.95  & 47.02  & 76.09  & 92.45  & 83.48 \\
    DAGMM  & 89.60  & 63.93  & 74.62  & 86.45  & 56.73  & 68.51  & 89.92  & 57.84  & 70.40  & 93.49  & 70.03  & 80.08 \\
    Deep-SVDD  & 91.92  & 76.63  & 83.58  & 89.93  & 56.02  & 69.04  & 80.42  & 84.45  & 82.39  & 95.41  & 86.49  & 90.73  \\
    LSTM-VAE  & 85.49  & 79.94  & 82.62  & 92.20  & 67.75  & 78.10  & 76.00  & 89.50  & 82.20  & 73.62  & 89.92  & 80.96  \\
    OmniAnomaly  & 89.02  & 86.37  & 87.67  & 92.49  & 81.99  & 86.92  & 81.42  & 84.30  & 82.83  & 88.39  & 74.46  & 80.83  \\
    THOC   & 88.45  & 90.97  & 89.69  & 92.06  & 89.34  & 90.68  & 83.94  & 86.36  & 85.13  & 88.14  & 90.99  & 89.54  \\
    InterFusion   & 81.28  & 92.70  & 86.62  & 89.77  & 88.52  & 89.14  & 80.59  & 85.58  & 83.01  & 83.61  & 83.45  & 83.52  \\
    AnomalyTrans   & 91.92  & 96.03  & 93.93  & 93.59  & \textcolor{blue}{\underline{99.41}} & 96.41  & 89.10  & 99.28  & 94.22  & 96.94  & 97.81  & 97.37  \\
    DCdetector   & 92.28  & \textcolor{blue}{\underline{97.21}}  & 94.68  & \textcolor{blue}{\underline{94.25}}  & 98.59  & 96.37  & 93.11  & \textcolor{blue}{\underline{99.77}}  & 96.33  & 97.14 & 98.74  & 97.94  \\
    MEMTO  & 92.07 & 96.76 & 94.36 & 93.76 & \textcolor{red}{\textbf{99.63}} & \textcolor{red}{\textbf{96.61}} & 94.18 & 97.54 & 95.83 & 97.46 & \textcolor{blue}{\underline{99.23}} & \textcolor{blue}{\underline{98.34}}  \\
    ModernTCN  & 83.94 & 85.93 &	84.92 &	93.17 &	57.69 &	71.26 &	91.83 &	95.98 &	93.86 &	\textcolor{red}{\textbf{98.09}} &	96.38 &	97.23  \\
    SensitiveHUE  & 90.86 & 79.13 &	84.59 &	88.90 &	60.62 &	72.09 &	\textcolor{red}{\textbf{96.93}} &	95.98 &	96.45 &	97.63 &	92.82 &	95.16  \\ 
    MtsCID  & \textcolor{blue}{\underline{93.37}} & 96.97 & \textcolor{blue}{\underline{95.13}} & 91.76 & 94.17 & 92.95 & 94.16 & \textcolor{red}{\textbf{99.82}} & \textcolor{blue}{\underline{96.91}} & 97.41 & 98.41 & 97.91 \\
    \midrule
    $\mathtt{GDformer}$  & \textcolor{red}{\textbf{93.70}} & \textcolor{red}{\textbf{98.07}} & \textcolor{red}{\textbf{95.83}} & \textcolor{red}{\textbf{95.55}} & 97.52 & \textcolor{blue}{\underline{96.52}} & \textcolor{blue}{\underline{96.28}} & \textcolor{red}{\textbf{99.82}} & \textcolor{red}{\textbf{98.02}} & \textcolor{blue}{\underline{97.97}} & \textcolor{red}{\textbf{99.52}} & \textcolor{red}{\textbf{98.74}} \\
    \bottomrule
    \end{tabular}%
  \label{results}%
\end{table*}%

\section{Experiments}
\label{Sec4}

\subsection{Experimental Details}
\label{appendix_setup}
\textbf{Datasets.}
We evaluate the anomaly detection performance on six widely used real-world MTS datasets. A summary of these datasets is elaborated in Table~\ref{details}, with the following descriptions.
\textbf{MSL} (Mars Science Laboratory dataset) is collected by NASA with 55 dimensions and shows the condition of the sensors and actuator data from the Mars rover \cite{10.1145/3219819.3219845}.
\textbf{SMAP} (Soil Moisture Active Passive dataset) records the soil samples of Mars with 25 dimensions \cite{10.1145/3219819.3219845}.
\textbf{SWaT} (Secure Water Treatment dataset) is collected from the critical infrastructure systems with 51 sensors \cite{7469060}.
 \textbf{PSM} (Pooled Server Metrics dataset) is collected from eBay server machines with 25 dimensions \cite{10.1145/3447548.3467174}.
 \textbf{GECCO} includes the recordings of the devices for detecting drinking data quality \cite{yang2023dcdetector,GECCO}.
 \textbf{ASD} (Application Server Dataset) contains 19 metrics for the status of servers \cite{li2021multivariate}.

\textbf{Implementation Settings.}
For the multi-entity MSL, SMAP, and ASD datasets, we concatenate the time series from all entities into a single and long multivariate time series and train a unified $\mathtt{GDformer}$ model on this combined dataset, following the standard protocol used in \cite{xu2022anomaly}.
The whole series is divided into multiple non-overlapped subsequences with $T=100$. 
The training data are divided into 80\% for training and 20\% for validation.
For $\mathtt{GDformer}$, we have $L=3$, the embedding dimension $D=512$, the number of cross-attention heads $H=8$. The mask ratio $\alpha$ is set to $5\%$.
$\delta$ indicates the top $\delta \%$ anomaly score is termed as the detection criterion.
The settings of the dictionary size $N$, the number of prototypes $P$, the loss trade-off parameter $\lambda$, and the threshold $\delta$ are shown in Table \ref{details}.
We investigate the effects of different parameter settings in Sec.~\ref{parameter-setting}.
We employ the ADAM \cite{DBLP:journals/corr/KingmaB14} with an initial learning rate of $10^{-4}$ to optimize model parameters.
The training process is continued for 10 epochs with the batch size of 64. 
All experiments are implemented in PyTorch \cite{NEURIPS2019_bdbca288} with a single NVIDIA GeForce RTX 3090 24GB GPU. The source code is available at the {anonymous repository}\footnote{https://anonymous.4open.science/r/GDformer-1F6C}.

\textbf{Baselines.} 
We evaluate the detection performance of $\mathtt{GDformer}$ with a broad collection of relevant methods, including the classic methods: OCSVM \cite{tax2004support} and IForest \cite{liu2008isolation}; density-estimation models: DAGMM \cite{zong2018deep}; clustering-based models: Deep-SVDD~\cite{ruff2018deep} and THOC~\cite{shen2020timeseries}; reconstruction-based models: LSTM-VAE~\cite{park2018multimodal}, OmniAnomaly \cite{su2019robust}, InterFusion \cite{li2021multivariate}, AnomalyTrans \cite{xu2022anomaly}, DCdetector \cite{yang2023dcdetector}, MEMTO~\cite{song2023memto}, ModernTCN~\cite{luo2024moderntcn}, SensitiveHUE~\cite{feng2024sensitivehue}, 
and MtsCID \cite{xie2025multivariate}.
The implementations of baselines are obtained in public repositories. For fair comparison, we adhere to the default parameters in the corresponding papers.

\textbf{Metrics.}
We adopt three groups of evaluation metrics, which can provide different evaluation views.
\begin{itemize}[leftmargin=*]
    \item The first group includes point-adjustment Precision (P), Recall (R), and F1-score (F1), which are widely adopted in \cite{xie2025multivariate,xu2022anomaly,yang2023dcdetector,song2023memto}. In this setting, once any outlier in a segment is detected, the whole anomaly segment is also identified.
    \item The second group includes Range-AUC-ROC (R-A-R) and Range-AUC-PR (R-A-PR), which are threshold-independent metrics and robust to misalignments with human labels \cite{paparrizos2022volume}. These metrics are also used in recent studies \cite{nam2024breaking,yang2023dcdetector}.
    \item The third group includes VUS-ROC (V-R) and VUS-PR (V-PR), which are adopted in recent studies \cite{yang2023dcdetector,xie2025multivariate,nam2024breaking}. These metrics are independent to the detection threshold and adaptive to continuous sequences \cite{paparrizos2022volume}.
\end{itemize}

\begin{table*}[!tbp]
  \centering
  \caption{ Anomaly detection performance comparisons in multiple metrics. \textcolor{red}{\textbf{Bold}}: the best. \textcolor{blue}{\underline{Underline}}: the second best.}
  \label{all_metrics}
  \begin{tabular}{ccccccccc}
    \toprule
    Dataset & {Metrics} & \makecell{AnomalyTrans } & \makecell{DCdetector } & \makecell{MEMTO } & \makecell{ModernTCN } & \makecell{SensitiveHUE } & \makecell{MtsCID } & \makecell{$\mathtt{GDformer}$ } \\
    \midrule
    \multirow{5}{*}{MSL} 
    & F1     & 93.93 & 94.68 & 94.36 & 84.92 & 84.59 & \textcolor{blue}{\underline{95.13}} & \textcolor{red}{\textbf{95.83}} \\
    & R-A-R  & \textcolor{blue}{\underline{90.17}} & 89.98 & 88.93 & 87.24 & 57.47 & 89.27 & \textcolor{red}{\textbf{90.89}} \\
    & R-A-PR & 87.96 & 87.87 & 86.93 & \textcolor{red}{\textbf{90.81}} & 58.06 & 87.12 & \textcolor{blue}{\underline{89.33}} \\
    & V-R    & 88.57 & 88.20 & 87.11 & 86.67 & 57.46 & 88.66 & \textcolor{red}{\textbf{90.22}} \\
    & V-PR   & 86.54 & 86.31 & 85.34 & \textcolor{red}{\textbf{90.39}} & 58.04 & 86.61 & \textcolor{blue}{\underline{88.78}} \\
    \midrule
    \multirow{5}{*}{SMAP} 
    & F1     & 96.41 & 96.37 & \textcolor{red}{\textbf{96.61}} & 71.26 & 72.09 & 92.95 & \textcolor{blue}{\underline{96.52}} \\
    & R-A-R  & 85.76 & \textcolor{blue}{\underline{95.87}} & 89.26 & 62.62 & 60.32 & 89.27 & \textcolor{red}{\textbf{96.81}} \\
    & R-A-PR & 85.76 & \textcolor{blue}{\underline{93.99}} & 88.84 & 64.27 & 58.38 & 87.12 & \textcolor{red}{\textbf{94.51}} \\
    & V-R    & 85.80 & \textcolor{blue}{\underline{94.78}} & 88.04 & 63.07 & 60.33 & \textcolor{blue}{\underline{88.66}} & \textcolor{red}{\textbf{96.23}} \\
    & V-PR   & 85.80 & \textcolor{blue}{\underline{93.03}} & 88.04 & 64.66 & 58.38 & 86.61 & \textcolor{red}{\textbf{94.01}} \\
    \midrule
    \multirow{5}{*}{SWaT} 
    & F1     & 94.22 & 96.33 & 95.83 & 93.86 & 96.45 & \textcolor{blue}{\underline{96.91}} & \textcolor{red}{\textbf{98.02}} \\
    & R-A-R  & 84.42 & \textcolor{blue}{\underline{96.61}} & 90.91 & 86.42 & 88.18 & 61.16 & \textcolor{red}{\textbf{98.37}} \\
    & R-A-PR & 79.91 & \textcolor{blue}{\underline{94.03}} & 72.34 & 85.87 & 88.77 & 65.91 & \textcolor{red}{\textbf{96.96}} \\
    & V-R    & 84.37 & \textcolor{blue}{\underline{96.81}} & 90.95 & 86.44 & 88.11 & 61.50 & \textcolor{red}{\textbf{98.09}} \\
    & V-PR   & 87.97 & \textcolor{blue}{\underline{94.21}} & 72.39 & 85.84 & 88.63 & 65.91 & \textcolor{red}{\textbf{96.72}} \\
    \midrule
    \multirow{5}{*}{PSM} 
    & F1     & 97.37 & 97.94 & \textcolor{blue}{\underline{98.34}} & 97.23 & 95.16 & 97.91 & \textcolor{red}{\textbf{98.74}} \\
    & R-A-R  & 89.38 & 86.66 & 90.08 & 87.24 & 80.29 & \textcolor{blue}{\underline{92.90}} & \textcolor{red}{\textbf{94.76}} \\
    & R-A-PR & 92.20 & 89.36 & 92.09 & 90.81 & 76.08 & \textcolor{blue}{\underline{94.17}} & \textcolor{red}{\textbf{95.50}} \\
    & V-R    & 87.81 & 82.38 & 89.81 & 86.67 & 79.66 & \textcolor{blue}{\underline{89.90}} & \textcolor{red}{\textbf{90.16}} \\
    & V-PR   & 91.07 & 86.14 & \textcolor{red}{\textbf{92.11}} & 90.39 & 75.59 & \textcolor{blue}{\underline{92.03}} & 91.95 \\
    \midrule
    \multirow{5}{*}{GECCO} 
    & F1     & 34.98 & 37.63 & 50.74 & 43.75 & 49.54 & \textcolor{blue}{\underline{57.27}} & \textcolor{red}{\textbf{59.20}} \\
    & R-A-R  & 60.60 & 60.03 & 56.07 & 53.14 & 56.64 & \textcolor{blue}{\underline{65.51}} & \textcolor{red}{\textbf{70.19}} \\
    & R-A-PR & 27.35 & 28.49 & 39.30 & 39.11 & 30.66 & \textcolor{red}{\textbf{39.87}} & \textcolor{blue}{\underline{39.40}} \\
    & V-R    & 61.66 & 59.17 & 56.55 & 53.03 & 57.23 & \textcolor{blue}{\underline{66.33}} & \textcolor{red}{\textbf{71.99}} \\
    & V-PR   & 10.14 & 27.63 & \textcolor{blue}{\underline{39.68}} & 38.08 & 33.34 & 35.90 & \textcolor{red}{\textbf{41.18}} \\
    \midrule
    \multirow{5}{*}{ASD} 
    & F1     & 84.76 & \textcolor{blue}{\underline{95.66}} & 95.38 & 92.77 & 85.62 & 95.44 & \textcolor{red}{\textbf{98.50}} \\
    & R-A-R  & 93.52 & 88.28 & 94.03 & \textcolor{red}{\textbf{95.12}} & 93.69 & 94.00 & \textcolor{blue}{\underline{94.21}} \\
    & R-A-PR & 87.29 & 84.49 & 90.78 & 90.80 & 84.05 & \textcolor{blue}{\underline{90.90}} & \textcolor{red}{\textbf{93.28}} \\
    & V-R    & 93.21 & 90.57 & \textcolor{blue}{\underline{94.26}} & 94.05 & 93.78 & 89.70 & \textcolor{red}{\textbf{95.19}} \\
    & V-PR   & 87.06 & 88.16 & \textcolor{blue}{\underline{90.96}} & 90.77 & 84.07 & 88.24 & \textcolor{red}{\textbf{91.00}} \\
    \midrule
    \multicolumn{2}{c}{$1^{\rm st}$ Count} & 0 & 0 & 2 & \textcolor{blue}{\underline{3}} & 0 & 1 & \textcolor{red}{\textbf{24}}\\
    \bottomrule
  \end{tabular}%
\end{table*}

\begin{table*}[!tbp]
  \centering
  \caption{Ablation results (F1-score) in $\mathtt{GDformer}$. The module is remarked with ``\ding{55}'', if we ablate it and ``\ding{51}'' otherwise. \textit{self-attention} and \textit{cross-attention} represent attention maps are from self-attention or dictionary-based cross attention mechanism.
  \textit{Recon} and \textit{Sim} represent the reconstruction error or the similarity-based criterion. \textbf{Bold}: the best.
  }
    \begin{tabular}{cccc|ccccc}
    \toprule
    Variant & $\mathcal{L}_c$    & $\mathcal{L}_s$    & Criterion & MSL   & SWaT  & PSM   & SMAP  & Avg \\
    \midrule
    \textbf{A.1} & self-attention & \ding{55}    & Recon & 88.94 & 94.29 & 93.72 & 76.76 & 88.43 \\
    \textbf{A.2} & self-attention & \ding{51}   & Recon & 88.61 & 94.04 & 92.84 & 72.59 & 87.02 \\
    \textbf{A.3} & self-attention & \ding{51}   & Sim & 92.33  & 93.35  & 97.70  & 82.24  & 91.41  \\
    \textbf{A.4} & cross-attention & \ding{51}   & Recon & 90.84  & 93.00  & 92.79  & 76.04  & 88.17  \\
    \textbf{A.5} & \ding{55}    & \ding{51}   & Sim & 95.21  & 94.65  & 98.03  & 75.69  & 90.90  \\
    \midrule
    \textbf{A.6} $\mathtt{GDformer}$ & cross-attention & \ding{51}   & Sim & \textbf{95.83} & \textbf{98.02} & \textbf{98.74} & \textbf{96.52} & \textbf{97.28} \\
    \bottomrule
    \end{tabular}%
  \label{ab1}%
\end{table*}%

\begin{table*}[!t]
  \centering
  \caption{Ablation results in $\mathcal{L}_s$. \textbf{Bold}: the best.}
    \begin{tabular}{c|ccc|ccc|ccc|ccc}
    \toprule
    Dataset & \multicolumn{3}{c|}{MSL} & \multicolumn{3}{c|}{SWaT} & \multicolumn{3}{c|}{PSM} & \multicolumn{3}{c}{SMAP} \\
    \midrule
    Metric & P     & R     & F1    & P     & R     & F1    & P     & R     & F1    & P     & R     & F1 \\
    \midrule
    \textbf{B.1} &  92.75  & 91.63  & 92.19 & 95.84  & 94.29  & 95.06  & 98.48  & 96.72  & 97.59  & \textbf{96.58} & 96.32  & 96.45  \\
    \textbf{B.2} &   92.34  & 81.77  & 86.73 & 95.78  & 92.97  & 94.36  & \textbf{98.63} & 95.69  & 97.14  & 94.72  & 60.88  & 74.12  \\
    \textbf{C.1} & 93.20  & 93.05  & 93.12  & 96.21  & 97.95  & 97.07  & 98.46  & 95.86  & 97.14  & 94.69  & 63.83  & 76.26  \\
    \textbf{C.2} & 93.35  & 93.46  & 93.40  & 95.94  & 94.26  & 95.10  & 98.57  & 96.08  & 97.31  & 94.27  & 57.40  & 71.36  \\
    \midrule
    $\mathtt{GDformer}$ & \textbf{93.70} & \textbf{98.07} & \textbf{95.83} & \textbf{96.28} & \textbf{99.82} & \textbf{98.02} & 97.97  & \textbf{99.52} & \textbf{98.74} & 95.55  & \textbf{97.52} & \textbf{96.52} \\
    \bottomrule
    \end{tabular}%
  \label{ab2}%
\end{table*}%

\begin{table*}[!tbp]
  \centering
  \caption{Transfer learning results of $\mathtt{GDformer}$. All metrics are organized in \%. \textcolor{red}{\textbf{Bold}}: the best. \textcolor{blue}{\underline{Underline}}: the second best. S: source datasets. T: target datasets. }
    \begin{tabular}{c|ccc|ccc|ccc|ccc}
    \toprule
    \multirow{2}[4]{*}{\diagbox{S}{T}} & \multicolumn{3}{c|}{PSM} & \multicolumn{3}{c|}{SMAP} & \multicolumn{3}{c|}{MSL} & \multicolumn{3}{c}{SWaT} \\
\cmidrule{2-13}          & P     & R     & F1    & P     & R     & F1    & P     & R     & F1    & P     & R     & F1 \\
    \midrule
    PSM   &  97.97 & \textcolor{red}{\textbf{99.52}} & \textcolor{red}{\textbf{98.74}} & 94.64 & 96.63 & 95.62 & \textcolor{blue}{\underline{92.78}} & \textcolor{red}{\textbf{98.07}} & \textcolor{blue}{\underline{95.35}} & 94.93 & \textcolor{red}{\textbf{99.82}} & 97.31 \\
    SMAP  & 97.97 & \textcolor{blue}{\underline{98.36}} & \textcolor{blue}{\underline{98.16}} & \textcolor{red}{\textbf{95.55}} & \textcolor{blue}{\underline{97.52}} & \textcolor{red}{\textbf{96.52}} & \textcolor{blue}{\underline{92.78}} & \textcolor{blue}{\underline{97.03}} & 94.86 & \textcolor{blue}{\underline{95.39}} & \textcolor{red}{\textbf{99.82}} & \textcolor{blue}{\underline{97.55}} \\
    MSL   & \textcolor{red}{\textbf{98.48}} & 97.56 & 98.02 & 94.38 & 96.57 & 95.46 & \textcolor{red}{\textbf{93.70}} & \textcolor{red}{\textbf{98.07}} & \textcolor{red}{\textbf{95.83}} & 94.16 & \textcolor{blue}{\underline{98.97}} & 96.50 \\
    SWaT  & \textcolor{blue}{\underline{98.36}} & 97.22 & 97.79 & \textcolor{blue}{\underline{94.69}} & \textcolor{red}{\textbf{97.87}} & \textcolor{blue}{\underline{96.25}} & 92.76 & \textcolor{red}{\textbf{98.07}} & 95.34 & \textcolor{red}{\textbf{96.28}} & \textcolor{red}{\textbf{99.82}} & \textcolor{red}{\textbf{98.02}} \\
    \bottomrule
    \end{tabular}%
  \label{transfer}%
\end{table*}%

\subsection{Main Results}
\label{main_results}
We first evaluate the anomaly detection performance using the point-adjustment metrics on four datasets, with comparison results presented in Table \ref{results}. Given that the most recent baselines, including AnomalyTrans, DCdetector, MEMTO, ModernTCN, SensitiveHUE, and MtsCID outperform other baselines, we conduct further comparison on two additional and more challenging datasets in multiple metrics, with comparison results presented in Table \ref{all_metrics}.
The numerical results in Table \ref{results} and \ref{all_metrics} demonstrate the SOTA performance of our proposed $\mathtt{GDformer}$.
Specifically, $\mathtt{GDformer}$ achieves the highest F1 scores in five out of six datasets and the second best in the remaining dataset. 
Furthermore, $\mathtt{GDformer}$ outperforms the baselines in the other two groups of threshold-independent metrics, with 24 top scores across all datasets. 
The notable performance gains of $\mathtt{GDformer}$ underscore the effectiveness of cultivating series-level representations, which breaks through the horizon limitation in subsequence isolation strategies.

\subsection{Model Analysis}


\subsubsection{Model Ablation.} 
\label{model_ablation}
The ablation results of loss function and detection criterion are shown in Table \ref{ab1}. 
The proposed similarity-based criterion brings \textbf{9.11\%} averaged F1-score improvements (from 88.17\% to 97.28\%), by comparing \textbf{A.4} and \textbf{A.6}. 
The dictionary-based cross attention mechanism can provide \textbf{5.87\%} averaged F1-score improvements (from 91.41\% to 97.28\%) by comparing \textbf{A.3} and \textbf{A.6}. 
In \textbf{A.5}, we employ dictionary-based cross attention mechanism to obtain the attention map but ablate $\mathcal{L}_c$ from Eq. (\ref{loss_total}). 
\textbf{A.5} achieves better performance compared with \textbf{A.1}, the pure Transformer, which validates our insight that similarity discrepancy might be a promising alternative to the reconstruction error in MTS anomaly detection.

\subsubsection{Similarity Evaluation Ablation.}
\label{similarity_ablation}
We conduct ablation studies on similarity discrepancy loss $\mathcal{L}_s$ and the numerical results are presented in Table \ref{ab2}. As is formulated in Eq. (\ref{loss_total}), $\mathcal{L}_s$ combines the cosine similarity in all $L$ layers. \textbf{B.1} (\textbf{B.2}) means only the first \textit{one} (\textit{two}) layer(s) is (are) considered in Eq. (\ref{loss_total}). \textbf{C.1} and \textbf{C.2} mean we adopt KL divergence and JD divergence respectively in Eq. (\ref{sim_eval}) to calculate the map-prototype similarity. 
As is shown in Table \ref{ab2}, all-layer combination achieves the best, due to the effective usage of multi-level features. 
Compared with \textbf{C.1} and \textbf{C.2}, $\mathtt{GDformer}$ is more possible to cultivate the diversity of prototypes, thereby effectively capturing the attention weights of normal-global representations. Therefore, $\mathtt{GDformer}$ outperforms \textbf{C.1} and \textbf{C.2}.

\subsubsection{Transferability}
\label{Transferability}
Given that one can train a unified model for cross-domain time series analysis~\cite{liang2024foundation},
we evaluate the transferability of the global dictionary and prototypes across datasets. These two objects are frozen and transferred to a target dataset, after they are optimized on a source dataset. The remaining parameters are then trained from scratch based on the target dataset.
As shown in Table~\ref{transfer}, compared with the original settings, the transfer performance has little F1-score reduction and is also comparative to the baselines.
It indicates that the normal points may have shared temporal representations across different datasets, thus laying the basis for the construction of MTS anomaly detection foundation model. 

\begin{figure}[!t]
	\centering
	\includegraphics[width=\columnwidth]{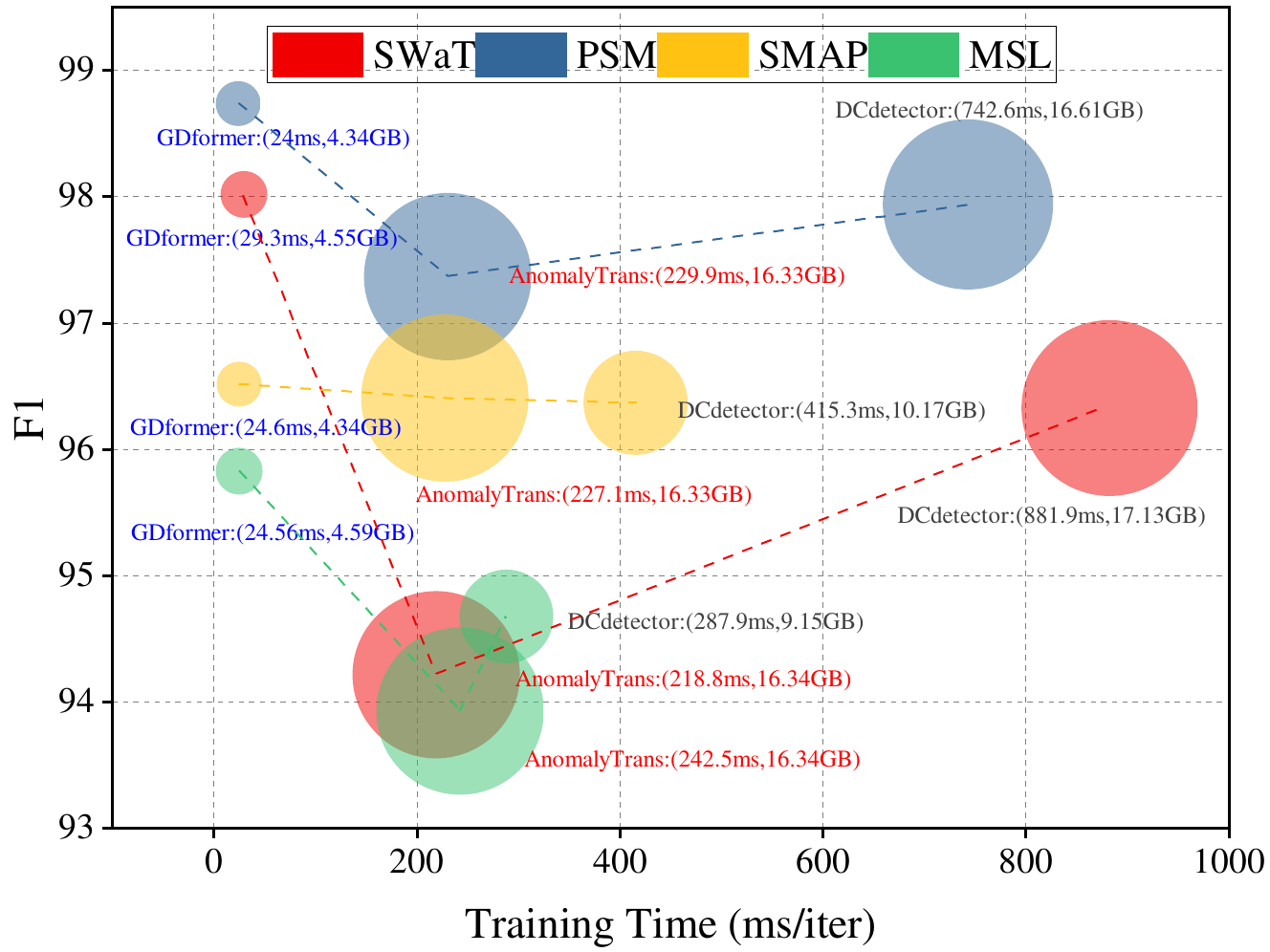}
	\caption{Model efficiency comparison in terms of training time, F1 score, and memory footprint. Larger bubble size indicates higher memory requirements.}
	\label{mod_eff}
\end{figure}

\begin{figure*}[!t]
	\centering
	\includegraphics[width=\textwidth]{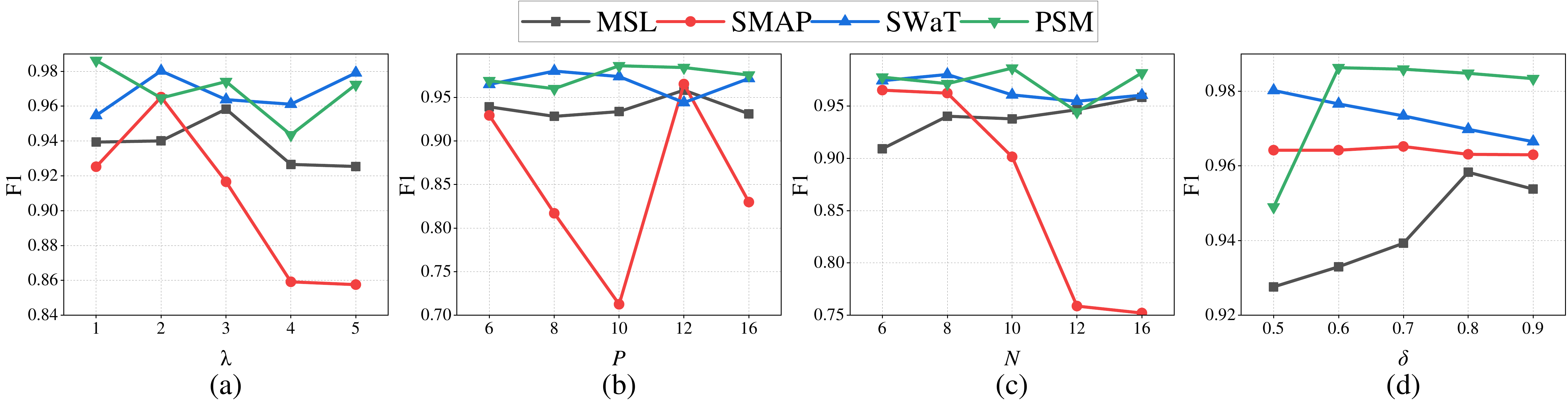}
	\caption{Parameter sensitivity analysis. (a) loss weight $\lambda$, (b) prototype size $P$, (c) dictionary size $N$, and (d) detection threshold $\delta$.}
	\label{param}
\end{figure*}

\begin{figure*}[!t]
	\centering
	\includegraphics[width=\textwidth]{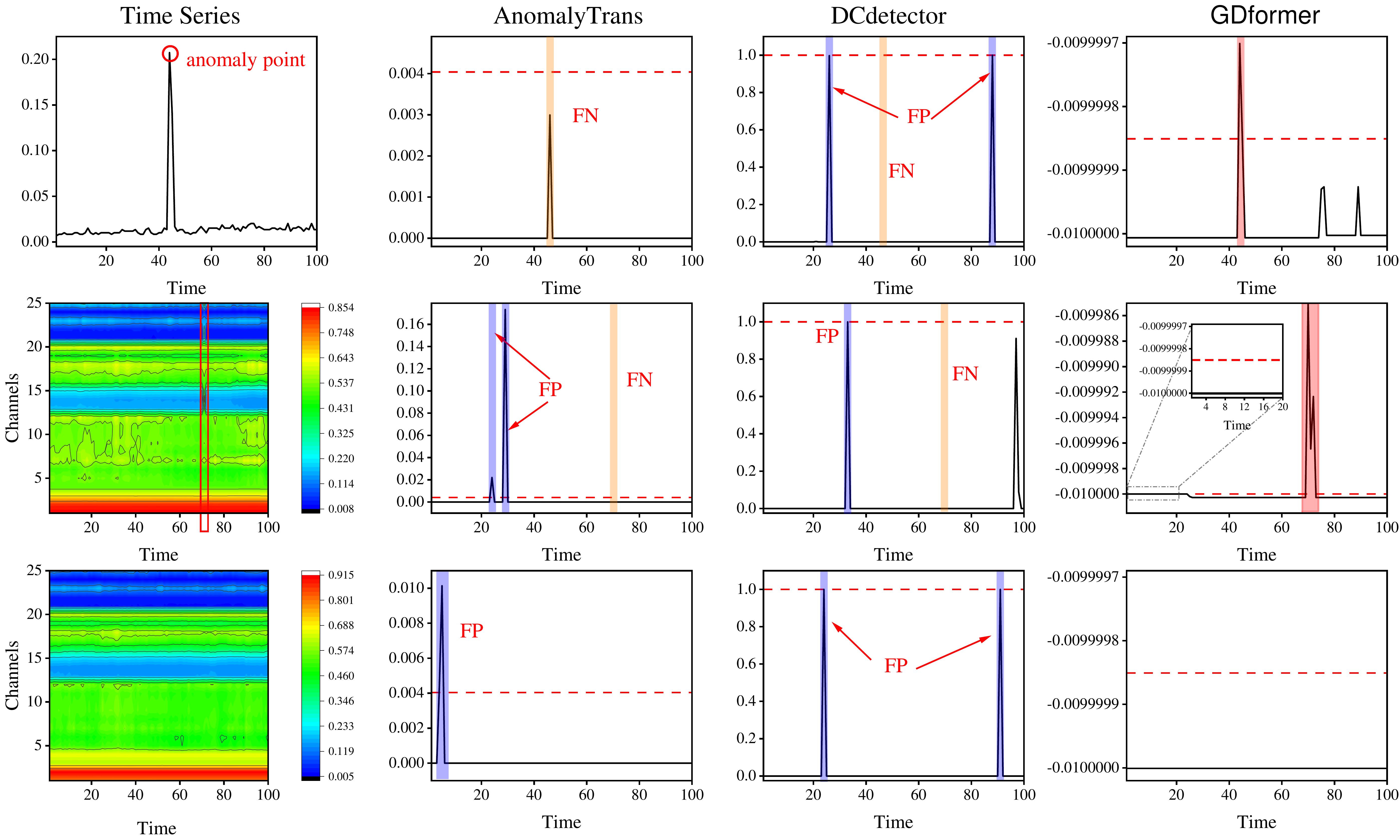}
	\caption{Detection results visualization of AnomalyTrans, DCdetector, and $\mathtt{GDformer}$ on PSM dataset. The point and segment anomalies are marked in \textcolor{red}{red circles} and \textcolor{red}{red boxes}. We plot the detection scores and the corresponding detection criteria (\textcolor{red}{red dashed lines}) for various methods. FP (false positive), FN (false negative) and true positive are highlighted in \textcolor{blue}{blue}, \textcolor{orange}{orange} and \textcolor{red}{red} respectively.}
	\label{case_study}
\end{figure*}
\subsubsection{Model Efficiency}
\label{model_efficiency}
We compare the model efficiency in terms of detection accuracy, training time, and memory footprint of the following methods: AnomalyTrans, DCdetector, and $\mathtt{GDformer}$.
As shown in Fig. \ref{mod_eff}, our proposed $\mathtt{GDformer}$ exceeds the other two Transformer-based methods consistently on four datasets. In self-attention module, the complexity can be formalized as $\mathcal{O}(T^2)$.
While in the devised coss-attention module, the complexity is $\mathcal{O}(TN)$, with $N$ much smaller than $T$ in our experiments.
Hence, the memory footprints of $\mathtt{GDformer}$ are lower than those of AnomalyTrans and DCdetector.
Moreover, the two baselines involve two-branch association modeling and two-stage optimization, which slows the training process.
In contrast, in $\mathtt{GDformer}$, the training time decreases significantly, with \textbf{88.8\%} and \textbf{94.7\%} averaged decline w.r.t AnomalyTrans and DCdetector.

\begin{figure*}[!t]
	\centering
	\includegraphics[width=\textwidth]{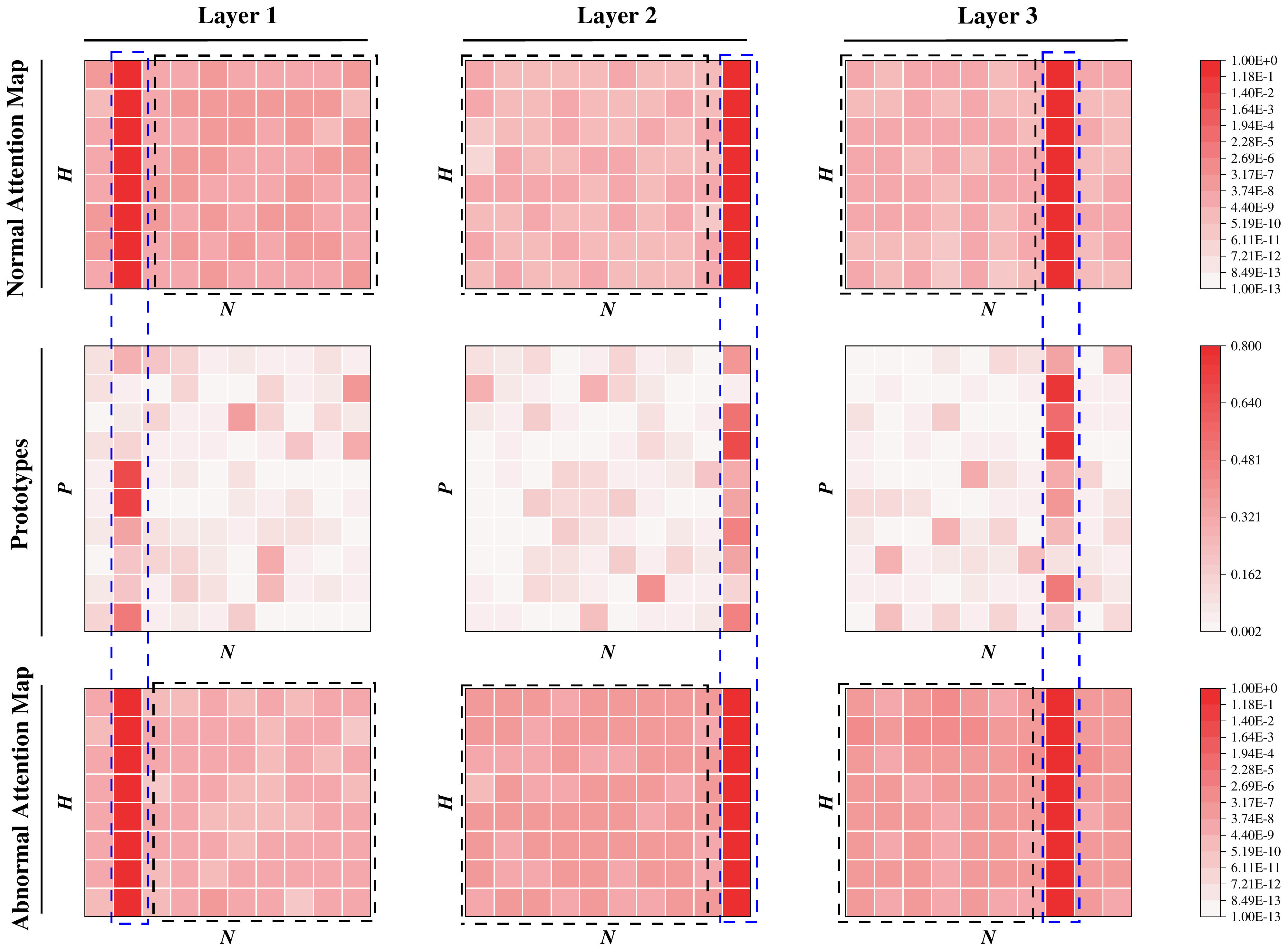}
	\caption{The showcase of prototypes (the second row) and attention maps on normal (the first row) and abnormal points (the third row). Each column corresponds to a layer. Each row has a unified color bar.}
	\label{attn_map}
\end{figure*}

\subsubsection{Sensitivity Investigation}
\label{parameter-setting}
Fig. \ref{param} shows the F1-score sensitivity on the four datasets. 
The loss weight $\lambda$ is adopted to balance the reconstruction loss and the distribution discrepancy loss. Higher values of $\lambda$ do not always guarantee higher F1-scores. We find that [1,3] may be an optimal range for all datasets.
Higher values of $P$ and $N$ have larger memory requirements. Less prototypes or key-value pairs may fail to capture the normal temporal patterns. On the other hand, more prototypes will carry redundant information, which subsequently leads to loose boundary.
We have the observation that how we design $\mathcal{L}_s$ have stronger effects on SMAP compared with the other three datasets, given F1-score on SMAP varies significantly with different settings of $\lambda$, $P$, and $N$.


\subsection{Case Study}
We showcase the detection results under the point and segment anomalies in Fig. \ref{case_study}.
We visualize one selected dimension for the point anomaly in the first row. 
A key observation is that AnomalyTrans and DCdetector fail to detect such point anomaly with the corresponding detection scores lower than anomaly criteria. Moreover, DCdetector even generates false positive cases.
The second row shows the segment anomalies. 
We visualize the MTS via a contour plot. It is clear that anomaly points range from 70 to 72. $\mathtt{GDformer}$ can consistently detect the segment anomalies. By contrast, the two baselines generate false cases.
In the third row, no anomalies exist in the input series, but the two baselines both yield false positive cases.
In general, AnomalyTrans and DCdetector focus on intra-subsequence point-wise association divergence, which is prone to be affected by the subsequence heterogeneity, and fail to promise compact series-level criteria, thus generating false cases.
$\mathtt{GDformer}$ can cultivate global representations with series-level knowledge and provide the unified criterion for any-position representations.

Fig.~\ref{attn_map} shows the prototypes and cross-attention scores of the normal and abnormal points.
A key observation is that the prototypical distribution of the association weights is unimodal.
Moreover, in the blue dashed box, the cross-attention scores of the normal and abnormal points are both in line with the above observation.
Therefore, directly adopting reconstruction errors as the detection criterion may lead to inferior accuracy.
However, the distribution of attention scores in the black dashed box vary on normal and abnormal points.
Hence, it naturally results in a distribution similarity-based criterion.


\section{Conclusion and Future Work}
\label{Sec5}
We proposes the global dictionary-enhanced Transformer model, $\mathtt{GDformer}$, to foster the learning of global representations shared by all normal points, which can solve the problem of limited horizons faced by the canonical Transformer.
Specifically, we renovate the self-attention mechanism into the dictionary-based cross-attention mechanism, where the Key and Value vectors in the global dictionary can learn the shared temporal representations. Moreover, the prototypes are introduced to capture the similarity distribution of normal points manifested by the cross-attention weights, which derives the similarity-based criterion. Extensive experiments validate the efficiency and effectiveness of $\mathtt{GDformer}$.

\textbf{Limitations and Future Work:}
The theoretical analysis of the functions of key-value pairs in the dictionary-based cross-attention mechanism will be conducted in future work. Moreover, given the transferability, we will explore the construction of foundation models for anomaly detection.



\appendix

\bibliographystyle{ACM-Reference-Format}
\bibliography{sample-base}

@String{Computing = "Computing" }

@String{Computer = "{IEEE} Computer" }

@String{Springer = "Springer-Verlag" }

@ArtifactSoftware{R,
    title = {R: A Language and Environment for Statistical Computing},
    author = {{R Core Team}},
    organization = {R Foundation for Statistical Computing},
    address = {Vienna, Austria},
    year = {2019},
    url = {https://www.R-project.org/},
}

@article{tax2004support,
  title={Support vector data description},
  author={Tax, David MJ and Duin, Robert PW},
  journal={Machine learning},
  volume={54},
  pages={45--66},
  year={2004},
  publisher={Springer}
}

@inproceedings{liu2008isolation,
  title={Isolation forest},
  author={Liu, Fei Tony and Ting, Kai Ming and Zhou, Zhi-Hua},
  booktitle={2008 eighth ieee international conference on data mining},
  pages={413--422},
  year={2008},
  organization={IEEE}
}

@inproceedings{breunig2000lof,
  title={LOF: identifying density-based local outliers},
  author={Breunig, Markus M and Kriegel, Hans-Peter and Ng, Raymond T and Sander, J{\"o}rg},
  booktitle={Proceedings of the 2000 ACM SIGMOD international conference on Management of data},
  pages={93--104},
  year={2000}
}

@article{yairi2017data,
  title={A data-driven health monitoring method for satellite housekeeping data based on probabilistic clustering and dimensionality reduction},
  author={Yairi, Takehisa and Takeishi, Naoya and Oda, Tetsuo and Nakajima, Yuta and Nishimura, Naoki and Takata, Noboru},
  journal={IEEE Transactions on Aerospace and Electronic Systems},
  volume={53},
  number={3},
  pages={1384--1401},
  year={2017},
  publisher={IEEE}
}

@inproceedings{zong2018deep,
  title={Deep autoencoding gaussian mixture model for unsupervised anomaly detection},
  author={Zong, Bo and Song, Qi and Min, Martin Renqiang and Cheng, Wei and Lumezanu, Cristian and Cho, Daeki and Chen, Haifeng},
  booktitle={International conference on learning representations},
  year={2018}
}

@inproceedings{ruff2018deep,
  title={Deep one-class classification},
  author={Ruff, Lukas and Vandermeulen, Robert and Goernitz, Nico and Deecke, Lucas and Siddiqui, Shoaib Ahmed and Binder, Alexander and M{\"u}ller, Emmanuel and Kloft, Marius},
  booktitle={International conference on machine learning},
  pages={4393--4402},
  year={2018},
  organization={PMLR}
}

@article{shen2020timeseries,
  title={Timeseries anomaly detection using temporal hierarchical one-class network},
  author={Shen, Lifeng and Li, Zhuocong and Kwok, James},
  journal={Advances in Neural Information Processing Systems},
  volume={33},
  pages={13016--13026},
  year={2020}
}

@inproceedings{shin2020itad,
  title={Itad: integrative tensor-based anomaly detection system for reducing false positives of satellite systems},
  author={Shin, Youjin and Lee, Sangyup and Tariq, Shahroz and Lee, Myeong Shin and Jung, Okchul and Chung, Daewon and Woo, Simon S},
  booktitle={Proceedings of the 29th ACM international conference on information \& knowledge management},
  pages={2733--2740},
  year={2020}
}

@inproceedings{hundman2018detecting,
  title={Detecting spacecraft anomalies using lstms and nonparametric dynamic thresholding},
  author={Hundman, Kyle and Constantinou, Valentino and Laporte, Christopher and Colwell, Ian and Soderstrom, Tom},
  booktitle={Proceedings of the 24th ACM SIGKDD international conference on knowledge discovery \& data mining},
  pages={387--395},
  year={2018}
}

@article{park2018multimodal,
  title={A multimodal anomaly detector for robot-assisted feeding using an lstm-based variational autoencoder},
  author={Park, Daehyung and Hoshi, Yuuna and Kemp, Charles C},
  journal={IEEE Robotics and Automation Letters},
  volume={3},
  number={3},
  pages={1544--1551},
  year={2018},
  publisher={IEEE}
}

@inproceedings{zhou2019beatgan,
  title={Beatgan: Anomalous rhythm detection using adversarially generated time series.},
  author={Zhou, Bin and Liu, Shenghua and Hooi, Bryan and Cheng, Xueqi and Ye, Jing},
  booktitle={IJCAI},
  volume={2019},
  pages={4433--4439},
  year={2019}
}

@inproceedings{su2019robust,
  title={Robust anomaly detection for multivariate time series through stochastic recurrent neural network},
  author={Su, Ya and Zhao, Youjian and Niu, Chenhao and Liu, Rong and Sun, Wei and Pei, Dan},
  booktitle={Proceedings of the 25th ACM SIGKDD international conference on knowledge discovery \& data mining},
  pages={2828--2837},
  year={2019}
}

@inproceedings{li2021multivariate,
  title={Multivariate time series anomaly detection and interpretation using hierarchical inter-metric and temporal embedding},
  author={Li, Zhihan and Zhao, Youjian and Han, Jiaqi and Su, Ya and Jiao, Rui and Wen, Xidao and Pei, Dan},
  booktitle={Proceedings of the 27th ACM SIGKDD conference on knowledge discovery \& data mining},
  pages={3220--3230},
  year={2021}
}

@inproceedings{
xu2022anomaly,
title={Anomaly Transformer: Time Series Anomaly Detection with Association Discrepancy},
author={Jiehui Xu and Haixu Wu and Jianmin Wang and Mingsheng Long},
booktitle={International Conference on Learning Representations},
year={2022},
url={https://openreview.net/forum?id=LzQQ89U1qm_}
}

@inproceedings{yang2023dcdetector,
  title={DCdetector: Dual Attention Contrastive Representation Learning for Time Series Anomaly Detection},
  author={Yiyuan Yang and Chaoli Zhang and Tian Zhou and Qingsong Wen and Liang Sun},
  booktitle={Proc. 29th ACM SIGKDD International Conference on Knowledge Discovery \& Data Mining (KDD 2023)},
  location = {Long Beach, CA},
  pages={3033–3045},
  year={2023}
}

@inproceedings{10.1145/3219819.3219845,
author = {Hundman, Kyle and Constantinou, Valentino and Laporte, Christopher and Colwell, Ian and Soderstrom, Tom},
title = {Detecting Spacecraft Anomalies Using LSTMs and Nonparametric Dynamic Thresholding},
year = {2018},
isbn = {9781450355520},
publisher = {Association for Computing Machinery},
address = {New York, NY, USA},
url = {https://doi.org/10.1145/3219819.3219845},
doi = {10.1145/3219819.3219845},
booktitle = {Proceedings of the 24th ACM SIGKDD International Conference on Knowledge Discovery \& Data Mining},
pages = {387–395},
numpages = {9},
location = {London, United Kingdom},
series = {KDD '18}
}

@INPROCEEDINGS{7469060,
  author={Mathur, Aditya P. and Tippenhauer, Nils Ole},
  booktitle={2016 International Workshop on Cyber-physical Systems for Smart Water Networks (CySWater)}, 
  title={SWaT: a water treatment testbed for research and training on ICS security}, 
  year={2016},
  volume={},
  number={},
  pages={31-36},
  doi={10.1109/CySWater.2016.7469060}}

@inproceedings{10.1145/3447548.3467174,
author = {Abdulaal, Ahmed and Liu, Zhuanghua and Lancewicki, Tomer},
title = {Practical Approach to Asynchronous Multivariate Time Series Anomaly Detection and Localization},
year = {2021},
isbn = {9781450383325},
publisher = {Association for Computing Machinery},
address = {New York, NY, USA},
url = {https://doi.org/10.1145/3447548.3467174},
doi = {10.1145/3447548.3467174},
booktitle = {Proceedings of the 27th ACM SIGKDD Conference on Knowledge Discovery \& Data Mining},
pages = {2485–2494},
numpages = {10},
location = {Virtual Event, Singapore},
series = {KDD '21}
}

@inproceedings{tang2002enhancing,
  title={Enhancing effectiveness of outlier detections for low density patterns},
  author={Tang, Jian and Chen, Zhixiang and Fu, Ada Wai-Chee and Cheung, David W},
  booktitle={Advances in Knowledge Discovery and Data Mining: 6th Pacific-Asia Conference, PAKDD 2002 Taipei, Taiwan, May 6--8, 2002 Proceedings 6},
  pages={535--548},
  year={2002},
  organization={Springer}
}

@article{yao2022kfreqgan,
  title={KfreqGAN: Unsupervised detection of sequence anomaly with adversarial learning and frequency domain information},
  author={Yao, Yueyue and Ma, Jianghong and Ye, Yunming},
  journal={Knowledge-Based Systems},
  volume={236},
  pages={107757},
  year={2022},
  publisher={Elsevier}
}

@inproceedings{li2021block,
  title={Block access pattern discovery via compressed full tensor transformer},
  author={Li, Xing and Shi, Qiquan and Hu, Gang and Chen, Lei and Mao, Hui and Yang, Yiyuan and Yuan, Mingxuan and Zeng, Jia and Cheng, Zhuo},
  booktitle={Proceedings of the 30th ACM International Conference on Information \& Knowledge Management},
  pages={957--966},
  year={2021}
}

@inproceedings{wen2022kddtimeseries,
title={Robust Time Series Analysis and Applications: An Industrial Perspective},
author={Wen, Qingsong and Yang, Linxiao and Zhou, Tian and Sun, Liang},
booktitle={Proceedings of the 28th ACM SIGKDD Conference on Knowledge Discovery \& Data Mining},
pages={4836--4837},
year={2022}
}

@misc{yang2023sgdpstreamgraphneuralnetwork,
      title={SGDP: A Stream-Graph Neural Network Based Data Prefetcher}, 
      author={Yiyuan Yang and Rongshang Li and Qiquan Shi and Xijun Li and Gang Hu and Xing Li and Mingxuan Yuan},
      year={2023},
      eprint={2304.03864},
      archivePrefix={arXiv},
      primaryClass={cs.OS},
      url={https://arxiv.org/abs/2304.03864}, 
}

@ARTICLE{9373923,
  author={Su, Ya and Zhao, Youjian and Sun, Ming and Zhang, Shenglin and Wen, Xidao and Zhang, Yongsu and Liu, Xian and Liu, Xiaozhou and Tang, Junliang and Wu, Wenfei and Pei, Dan},
  journal={IEEE Transactions on Computers}, 
  title={Detecting Outlier Machine Instances Through Gaussian Mixture Variational Autoencoder With One Dimensional CNN}, 
  year={2022},
  volume={71},
  number={4},
  pages={892-905},
  doi={10.1109/TC.2021.3065073}}

@misc{zhang2018deepneuralnetworkunsupervised,
      title={A Deep Neural Network for Unsupervised Anomaly Detection and Diagnosis in Multivariate Time Series Data}, 
      author={Chuxu Zhang and Dongjin Song and Yuncong Chen and Xinyang Feng and Cristian Lumezanu and Wei Cheng and Jingchao Ni and Bo Zong and Haifeng Chen and Nitesh V. Chawla},
      year={2018},
      eprint={1811.08055},
      archivePrefix={arXiv},
      primaryClass={cs.LG},
      url={https://arxiv.org/abs/1811.08055}, 
}

@misc{zhao2020multivariatetimeseriesanomalydetection,
      title={Multivariate Time-series Anomaly Detection via Graph Attention Network}, 
      author={Hang Zhao and Yujing Wang and Juanyong Duan and Congrui Huang and Defu Cao and Yunhai Tong and Bixiong Xu and Jing Bai and Jie Tong and Qi Zhang},
      year={2020},
      eprint={2009.02040},
      archivePrefix={arXiv},
      primaryClass={cs.LG},
      url={https://arxiv.org/abs/2009.02040}, 
}

@inproceedings{Zhang_2022, 
   title={TFAD: A Decomposition Time Series Anomaly Detection Architecture with Time-Frequency Analysis},
   url={http://dx.doi.org/10.1145/3511808.3557470},
   DOI={10.1145/3511808.3557470},
   booktitle={Proceedings of the 31st ACM International Conference on Information \& Knowledge Management},
   publisher={ACM},
   author={Zhang, Chaoli and Zhou, Tian and Wen, Qingsong and Sun, Liang},
   year={2022},
   month=oct, pages={2497–2507},
   collection={CIKM ’22} }

@inproceedings{10.5555/3618408.3619209,
author = {Li, Yuxin and Chen, Wenchao and Chen, Bo and Wang, Dongsheng and Tian, Long and Zhou, Mingyuan},
title = {Prototype-oriented unsupervised anomaly detection for multivariate time series},
year = {2023},
publisher = {JMLR.org},
booktitle = {Proceedings of the 40th International Conference on Machine Learning},
articleno = {801},
numpages = {18},
location = {Honolulu, Hawaii, USA},
series = {ICML'23}
}

@inproceedings{
kim2022reversible,
title={Reversible Instance Normalization for Accurate Time-Series Forecasting against Distribution Shift},
author={Taesung Kim and Jinhee Kim and Yunwon Tae and Cheonbok Park and Jang-Ho Choi and Jaegul Choo},
booktitle={International Conference on Learning Representations},
year={2022},
url={https://openreview.net/forum?id=cGDAkQo1C0p}
}

@InProceedings{Ulyanov_2017_CVPR,
author = {Ulyanov, Dmitry and Vedaldi, Andrea and Lempitsky, Victor},
title = {Improved Texture Networks: Maximizing Quality and Diversity in Feed-Forward Stylization and Texture Synthesis},
booktitle = {Proceedings of the IEEE Conference on Computer Vision and Pattern Recognition (CVPR)},
month = {July},
year = {2017}
}

@inproceedings{NIPS2017_3f5ee243,
 author = {Vaswani, Ashish and Shazeer, Noam and Parmar, Niki and Uszkoreit, Jakob and Jones, Llion and Gomez, Aidan N and Kaiser, \L ukasz and Polosukhin, Illia},
 booktitle = {Advances in Neural Information Processing Systems},
 editor = {I. Guyon and U. Von Luxburg and S. Bengio and H. Wallach and R. Fergus and S. Vishwanathan and R. Garnett},
 pages = {},
 publisher = {Curran Associates, Inc.},
 title = {Attention is All you Need},
 url = {https://proceedings.neurips.cc/paper_files/paper/2017/file/3f5ee243547dee91fbd053c1c4a845aa-Paper.pdf},
 volume = {30},
 year = {2017}
}

@ARTICLE{6832827,
  author={van Erven, Tim and Harremos, Peter},
  journal={IEEE Transactions on Information Theory}, 
  title={Rényi Divergence and Kullback-Leibler Divergence}, 
  year={2014},
  volume={60},
  number={7},
  pages={3797-3820},
  doi={10.1109/TIT.2014.2320500}}

@INPROCEEDINGS{1365067,
  author={Fuglede, B. and Topsoe, F.},
  booktitle={International Symposium onInformation Theory, 2004. ISIT 2004. Proceedings.}, 
  title={Jensen-Shannon divergence and Hilbert space embedding}, 
  year={2004},
  volume={},
  number={},
  pages={31-},
  doi={10.1109/ISIT.2004.1365067}}

@inproceedings{DBLP:journals/corr/KingmaB14,
  author       = {Diederik P. Kingma and
                  Jimmy Ba},
  editor       = {Yoshua Bengio and
                  Yann LeCun},
  title        = {Adam: {A} Method for Stochastic Optimization},
  booktitle    = {3rd International Conference on Learning Representations, {ICLR} 2015,
                  San Diego, CA, USA, May 7-9, 2015, Conference Track Proceedings},
  year         = {2015},
  url          = {http://arxiv.org/abs/1412.6980},
  bibsource    = {dblp computer science bibliography, https://dblp.org}
}

@inproceedings{NEURIPS2019_bdbca288,
 author = {Paszke, Adam and Gross, Sam and Massa, Francisco and Lerer, Adam and Bradbury, James and Chanan, Gregory and Killeen, Trevor and Lin, Zeming and Gimelshein, Natalia and Antiga, Luca and Desmaison, Alban and Kopf, Andreas and Yang, Edward and DeVito, Zachary and Raison, Martin and Tejani, Alykhan and Chilamkurthy, Sasank and Steiner, Benoit and Fang, Lu and Bai, Junjie and Chintala, Soumith},
 booktitle = {Advances in Neural Information Processing Systems},
 editor = {H. Wallach and H. Larochelle and A. Beygelzimer and F. d\textquotesingle Alch\'{e}-Buc and E. Fox and R. Garnett},
 pages = {},
 publisher = {Curran Associates, Inc.},
 title = {PyTorch: An Imperative Style, High-Performance Deep Learning Library},
 url = {https://proceedings.neurips.cc/paper_files/paper/2019/file/bdbca288fee7f92f2bfa9f7012727740-Paper.pdf},
 volume = {32},
 year = {2019}
}

@inproceedings{liang2024foundation,
  title={Foundation models for time series analysis: A tutorial and survey},
  author={Liang, Yuxuan and Wen, Haomin and Nie, Yuqi and Jiang, Yushan and Jin, Ming and Song, Dongjin and Pan, Shirui and Wen, Qingsong},
  booktitle={Proceedings of the 30th ACM SIGKDD conference on knowledge discovery and data mining},
  pages={6555--6565},
  year={2024}
}

@article{song2023memto,
  title={Memto: Memory-guided transformer for multivariate time series anomaly detection},
  author={Song, Junho and Kim, Keonwoo and Oh, Jeonglyul and Cho, Sungzoon},
  journal={Advances in Neural Information Processing Systems},
  volume={36},
  pages={57947--57963},
  year={2023}
}

@inproceedings{xiao2023imputation,
  title={Imputation-based time-series anomaly detection with conditional weight-incremental diffusion models},
  author={Xiao, Chunjing and Gou, Zehua and Tai, Wenxin and Zhang, Kunpeng and Zhou, Fan},
  booktitle={Proceedings of the 29th ACM SIGKDD conference on knowledge discovery and data mining},
  pages={2742--2751},
  year={2023}
}

@article{he2025graph,
  title={Graph-enhanced anomaly detection framework in multivariate time series using Graph Attention and Enhanced Generative Adversarial Networks},
  author={He, Yue and Chen, Xiaoliang and Miao, Duoqian and Zhang, Hongyun and Qin, Xiaolin and Du, Shangyi and Lu, Peng},
  journal={Expert Systems with Applications},
  volume={271},
  pages={126667},
  year={2025},
  publisher={Elsevier}
}

@article{paparrizos2022volume,
  title={Volume under the surface: a new accuracy evaluation measure for time-series anomaly detection},
  author={Paparrizos, John and Boniol, Paul and Palpanas, Themis and Tsay, Ruey S and Elmore, Aaron and Franklin, Michael J},
  journal={Proceedings of the VLDB Endowment},
  volume={15},
  number={11},
  pages={2774--2787},
  year={2022},
  publisher={VLDB Endowment}
}

@inproceedings{qiu2025tab,
title      = {{TAB}: Unified Benchmarking of Time Series Anomaly Detection Methods},
author     = {Xiangfei Qiu and Zhe Li and Wanghui Qiu and Shiyan Hu and Lekui Zhou and Xingjian Wu and Zhengyu Li and Chenjuan Guo and Aoying Zhou and Zhenli Sheng and Jilin Hu and Christian S. Jensen and Bin Yang},
booktitle  = {Proc. {VLDB} Endow.},
year       = {2025}
}

@inproceedings{xie2025multivariate,
  title={Multivariate Time Series Anomaly Detection by Capturing Coarse-Grained Intra-and Inter-Variate Dependencies},
  author={Xie, Yongzheng and Zhang, Hongyu and Babar, Muhammad Ali},
  booktitle={Proceedings of the ACM on Web Conference 2025},
  pages={697--705},
  year={2025}
}

@inproceedings{yu2025merlin,
  title={Merlin: Multi-View Representation Learning for Robust Multivariate Time Series Forecasting with Unfixed Missing Rates},
  author={Yu, Chengqing and Wang, Fei and Yang, Chuanguang and Shao, Zezhi and Sun, Tao and Qian, Tangwen and Wei, Wei and An, Zhulin and Xu, Yongjun},
  booktitle={Proceedings of the 31st ACM SIGKDD Conference on Knowledge Discovery and Data Mining V. 2},
  pages={3633--3644},
  year={2025}
}

@inproceedings{wu2024multi,
  title={Multi-view Self-Supervised Contrastive Learning for Multivariate Time Series},
  author={Wu, Yuhan and Meng, Xiyu and He, Yang and Zhang, Junru and Zhang, Haowen and Dong, Yabo and Lu, Dongming},
  booktitle={Proceedings of the 32nd ACM International Conference on Multimedia},
  pages={9582--9590},
  year={2024}
}

@inproceedings{liu2024wftnet,
  title={Wftnet: Exploiting global and local periodicity in long-term time series forecasting},
  author={Liu, Peiyuan and Wu, Beiliang and Li, Naiqi and Dai, Tao and Lei, Fengmao and Bao, Jigang and Jiang, Yong and Xia, Shu-Tao},
  booktitle={ICASSP 2024-2024 IEEE International Conference on Acoustics, Speech and Signal Processing (ICASSP)},
  pages={5960--5964},
  year={2024},
  organization={IEEE}
}

@misc{sánchezferrera2025reviewselfsupervisedlearningtime,
      title={A Review on Self-Supervised Learning for Time Series Anomaly Detection: Recent Advances and Open Challenges}, 
      author={Aitor Sánchez-Ferrera and Borja Calvo and Jose A. Lozano},
      year={2025},
      eprint={2501.15196},
      archivePrefix={arXiv},
      primaryClass={stat.ML},
      url={https://arxiv.org/abs/2501.15196}, 
}

@article{lu2008network,
  title={Network anomaly detection based on wavelet analysis},
  author={Lu, Wei and Ghorbani, Ali A},
  journal={EURASIP Journal on Advances in Signal processing},
  volume={2009},
  number={1},
  pages={837601},
  year={2008},
  publisher={Springer}
}

@inproceedings{mahimkar2011rapid,
  title={Rapid detection of maintenance induced changes in service performance},
  author={Mahimkar, Ajay and Ge, Zihui and Wang, Jia and Yates, Jennifer and Zhang, Yin and Emmons, Joanne and Huntley, Brian and Stockert, Mark},
  booktitle={Proceedings of the Seventh COnference on Emerging Networking EXperiments and Technologies},
  pages={1--12},
  year={2011}
}

@inproceedings{rasheed2009fourier,
  title={Fourier transform based spatial outlier mining},
  author={Rasheed, Faraz and Peng, Peter and Alhajj, Reda and Rokne, Jon},
  booktitle={International Conference on Intelligent Data Engineering and Automated Learning},
  pages={317--324},
  year={2009},
  organization={Springer}
}

@inproceedings{siffer2017anomaly,
  title={Anomaly detection in streams with extreme value theory},
  author={Siffer, Alban and Fouque, Pierre-Alain and Termier, Alexandre and Largouet, Christine},
  booktitle={Proceedings of the 23rd ACM SIGKDD international conference on knowledge discovery and data mining},
  pages={1067--1075},
  year={2017}
}

@inproceedings{khraisat2018anomaly,
  title={An anomaly intrusion detection system using C5 decision tree classifier},
  author={Khraisat, Ansam and Gondal, Iqbal and Vamplew, Peter},
  booktitle={Pacific-Asia Conference on Knowledge Discovery and Data Mining},
  pages={149--155},
  year={2018},
  organization={Springer}
}

@inproceedings{amer2013enhancing,
  title={Enhancing one-class support vector machines for unsupervised anomaly detection},
  author={Amer, Mennatallah and Goldstein, Markus and Abdennadher, Slim},
  booktitle={Proceedings of the ACM SIGKDD workshop on outlier detection and description},
  pages={8--15},
  year={2013}
}

@inproceedings{hu2003robust,
  title={Robust anomaly detection using support vector machines},
  author={Hu, Wenjie and Liao, Yihua and Vemuri, V Rao},
  booktitle={Proceedings of the international conference on machine learning},
  volume={6},
  pages={5},
  year={2003},
  organization={Citeseer University Park, PA, USA}
}

@inproceedings{luo2024moderntcn,
  title={Moderntcn: A modern pure convolution structure for general time series analysis},
  author={Luo, Donghao and Wang, Xue},
  booktitle={The twelfth international conference on learning representations},
  pages={1--43},
  year={2024}
}

@inproceedings{feng2024sensitivehue,
  title={Sensitivehue: Multivariate time series anomaly detection by enhancing the sensitivity to normal patterns},
  author={Feng, Yuye and Zhang, Wei and Fu, Yao and Jiang, Weihao and Zhu, Jiang and Ren, Wenqi},
  booktitle={Proceedings of the 30th ACM SIGKDD Conference on knowledge discovery and data mining},
  pages={782--793},
  year={2024}
}

@inproceedings{GECCO,
    author = {Moritz, Steffen and Rehbach, Frederik and Chandrasekaran, Sowmya and Rebolledo, Margarita and Bartz-Beielstein, Thomas },
    title = {GECCO Industrial Challenge 2018 Dataset: A water quality dataset for the 'Internet of Things: Online Anomaly Detection for Drinking Water Quality' competition},
    booktitle = {the Genetic and Evolutionary Computation Conference},
    year = {2018},
}

@inproceedings{nam2024breaking,
  title={Breaking the time-frequency granularity discrepancy in time-series anomaly detection},
  author={Nam, Youngeun and Yoon, Susik and Shin, Yooju and Bae, Minyoung and Song, Hwanjun and Lee, Jae-Gil and Lee, Byung Suk},
  booktitle={Proceedings of the ACM Web Conference 2024},
  pages={4204--4215},
  year={2024}
}


\end{document}